\documentclass[pmlr,twocolumn,10pt]{jmlr}

\mlhtrack{proceedings}

\newif\iffinal
\finaltrue  

\iffinal
    \jmlrproceedings{}{arXiv preprint}
    \jmlryear{2026}
    \jmlrworkshop{}
\else
    \jmlrproceedings{}{}
    \jmlrworkshop{}
\fi

\graphicspath{{./}{ML4H/}}
\usepackage{xurl}
\DeclareUrlCommand\path{\urlstyle{tt}%
  \def\UrlBreaks{\do\/\do\_\do\.\do\-}%
  \def\UrlBigBreaks{}}
\usepackage{microtype}
\usepackage{booktabs}
\usepackage{siunitx}
\SetSymbolFont{operators}{bold}{OT1}{cmr}{b}{n}
\usepackage{array}
\usepackage{longtable}
\usepackage{enumitem}
\usepackage{placeins}
\iffinal\else
\usepackage[switch]{lineno}
\fi
\usepackage[english]{babel}
\usepackage[autostyle]{csquotes}

\makeatletter
\AtBeginDocument{%
\long\def\@makecaption#1#2{%
   \vskip 10pt
   \setbox\@tempboxa\hbox{#1. #2}%
   \ifdim \wd\@tempboxa >\hsize
       \begin{list}{#1.}{%
       \settowidth{\labelwidth}{#1.}
       \setlength{\leftmargin}{\labelwidth}
       \addtolength{\leftmargin}{\labelsep}
        }\item #2 \end{list}\par
     \else
       \hbox to\hsize{\hfil\box\@tempboxa\hfil}
   \fi}%
\def\LT@makecaption#1#2#3{%
  \LT@mcol\LT@cols c{\hbox to\z@{\hss\parbox[t]\LTcapwidth{%
    \normalsize\normalfont
    \sbox\@tempboxa{#1{#2. }#3}%
    \ifdim\wd\@tempboxa>\hsize
      \ifx#1\@gobble
        #3%
      \else
        \begin{list}{#2.}{%
          \settowidth{\labelwidth}{#2.}%
          \setlength{\leftmargin}{\labelwidth}%
          \addtolength{\leftmargin}{\labelsep}%
          \setlength{\topsep}{0pt}%
          \setlength{\partopsep}{0pt}%
          \setlength{\itemsep}{0pt}%
          \setlength{\parsep}{0pt}%
        }\item #3\end{list}%
      \fi
    \else
      \hbox to\hsize{\hfil\box\@tempboxa\hfil}%
    \fi
    \endgraf\vskip 0.75\baselineskip}%
  \hss}}}%
}
\makeatother

\title[Representation Before Training]{Representation Before Training: A Practical Benchmark\titletag{\\} for Generative Medical Event Model Tokenization}

\iffinal
\newcommand{\authorcell}[3]{\begin{tabular}[t]{@{}c@{}}
  \Name{#1}\nametag{#2}\\[0pt]
  {\let\hfill\relax\Email{#3}}\end{tabular}}
\author{%
\noindent\hfill
\authorcell{Inhyeok Lee}{$^{1,2}$}{ihlee@uchicago.edu}\hspace{1.8em}%
\authorcell{Luke Solo}{$^{1}$}{lsolo@uchicago.edu}\hspace{1.8em}%
\authorcell{Michael C. Burkhart}{$^{1}$}{burkh4rt@uchicago.edu}\hfill\null\AND
\noindent\hfill
\authorcell{Bashar Ramadan}{$^{1}$}{basharramadan@uchicago.edu}\hspace{1.8em}%
\authorcell{Sahil Sethi}{$^{1}$}{sethis@uchicago.edu}\hspace{1.8em}%
\authorcell{Sarah Jabbour}{$^{3}$}{sarah.jabbour@chicagobooth.edu}\hfill\null\AND
\noindent\hfill
\authorcell{William F. Parker}{$^{1}$}{wparker@uchicago.edu}\hspace{1.8em}%
\authorcell{Brett K. Beaulieu-Jones}{$^{1,2}$}{beaulieujones@uchicago.edu}\hfill\null\\[8pt]
{\centering\addr
$^{1}$\,Center for Computational Medicine \textup{\&} Clinical AI, Section of Biomedical Data Science,\\
Department of Medicine, University of Chicago, Chicago, IL, USA\\
$^{2}$\,Committee on Genetics, Genomics, and Systems Biology, University of Chicago, Chicago, IL, USA\\
$^{3}$\,Booth School of Business, University of Chicago, Chicago, IL, USA\endgraf}
}
\else
\author{Anonymous Author(s)}
\linenumbers
\fi

\begin{document}
\raggedbottom

\maketitle

\begin{abstract}
Generative medical event models use tokenized sequences of patient timelines as input, but practical guidance on the many decisions around tokenization is limited because most studies keep the tokenizer fixed while scaling architecture or data. We benchmark quantization granularity, reference-range anchoring, code--value fusion, numeric and temporal encodings, and schema harmonization across 156 models. Using both Llama-3.2-style and Qwen3-style architectures, all models were trained on full MIMIC-IV hospitalizations from three initialization seeds under a shared training recipe, and linear probes evaluated frozen 24-hour representations across eight binary and regression outcome families during hours 24--48. Fused tokens pairing codes with value deciles produced statistically significant gains in six of eight families over unfused deciles, including all four regression families ($+0.032$ to $+0.114$ Spearman correlation) and vital-signs classification ($+0.033$ AUROC), with 38 of 48 comparisons significant across all six quantization schemes (seven of eight families). Next-token training organized unfused and fused centile embeddings into ordered trajectories and made clinical reference ranges linearly decodable in unfused centiles (93--97\% accuracy). However, neither reference-range anchoring nor finer quantization consistently improved 24-hour probes, while xVal variants showed statistically significant deficits across all eight families. Event order and admission-relative rotary position embeddings (RoPE) shortened inputs while marginally raising all eight family means over explicit time tokens (five and four statistically significant). Mapping native input to the Common Longitudinal Intensive Care Unit Data Format (CLIF) cut training tokens by 71.4\% and vocabulary size by 97.8\%, with statistically significant gains in four of eight families and decreases in two. These findings show that compact, expert-curated tokenizations can outperform more granular and continuous alternatives under a shared training recipe.
\end{abstract}

\begin{keywords}
Electronic health records, Generative medical event models, Tokenization, Quantization, Numeric and temporal encoding, Schema harmonization
\end{keywords}

\ifmlhneedsstatements
\paragraph*{Data and Code Availability}
We use the MIMIC-IV dataset \citep{Johnson23MIMICIV}, version 3.1, distributed through PhysioNet \citep{Johnson24MIMICIV31}, subject to data use requirements. For MIMIC-to-CLIF schema remapping, we use the CLIF~2.1.0 conversion of MIMIC-IV \citep{Liao26MIMICIVExtCLIF}.
\iffinal
All source code to reproduce our experiments is available at \href{https://github.com/bbj-lab/input-representation-benchmark}{\texttt{bbj-lab/\allowbreak input-representation-benchmark}} and \href{https://github.com/bbj-lab/fms-ehrs/tree/ml4h-2026}{\texttt{bbj-lab/\allowbreak fms-ehrs}} (tag \texttt{ml4h-2026}).
\else
{\sloppy
Anonymized copies of both codebases are available at
\url{https://anonymous.4open.science/r/representation-before-training-142A/}.
\par}
\fi

\paragraph*{Institutional Review Board (IRB)}
We used the MIMIC-IV database in accordance with PhysioNet terms, so no additional IRB approval was required.
\fi

\section{Introduction}
\label{sec:intro}
Electronic health records (EHRs) capture patient trajectories as a sequence of clinical events, measurements, and interventions. Generative medical event models learn from these sequences through next-token prediction, yielding representations that can be used for downstream prediction tasks, yet before training begins there are many decisions regarding how clinical data are turned into tokenized sequences. Numeric values, timing information, and clinical schemas all change how clinical information is expressed when training the transformer, and information lost during tokenization cannot be recovered downstream, even though most benchmarks vary architecture, objective, or scale on a fixed tokenizer. Potassium values of 5.5 and 7.0~mEq/L, for instance, both exceed the reference range with different severities of hyperkalemia and cardiac risk \citep{Weiss17Hyperkalemia}, yet decile-based quantization is likely to group these nuanced values into the same bin.

\begin{figure*}[t]
	\centering
		\includegraphics[width=\textwidth]{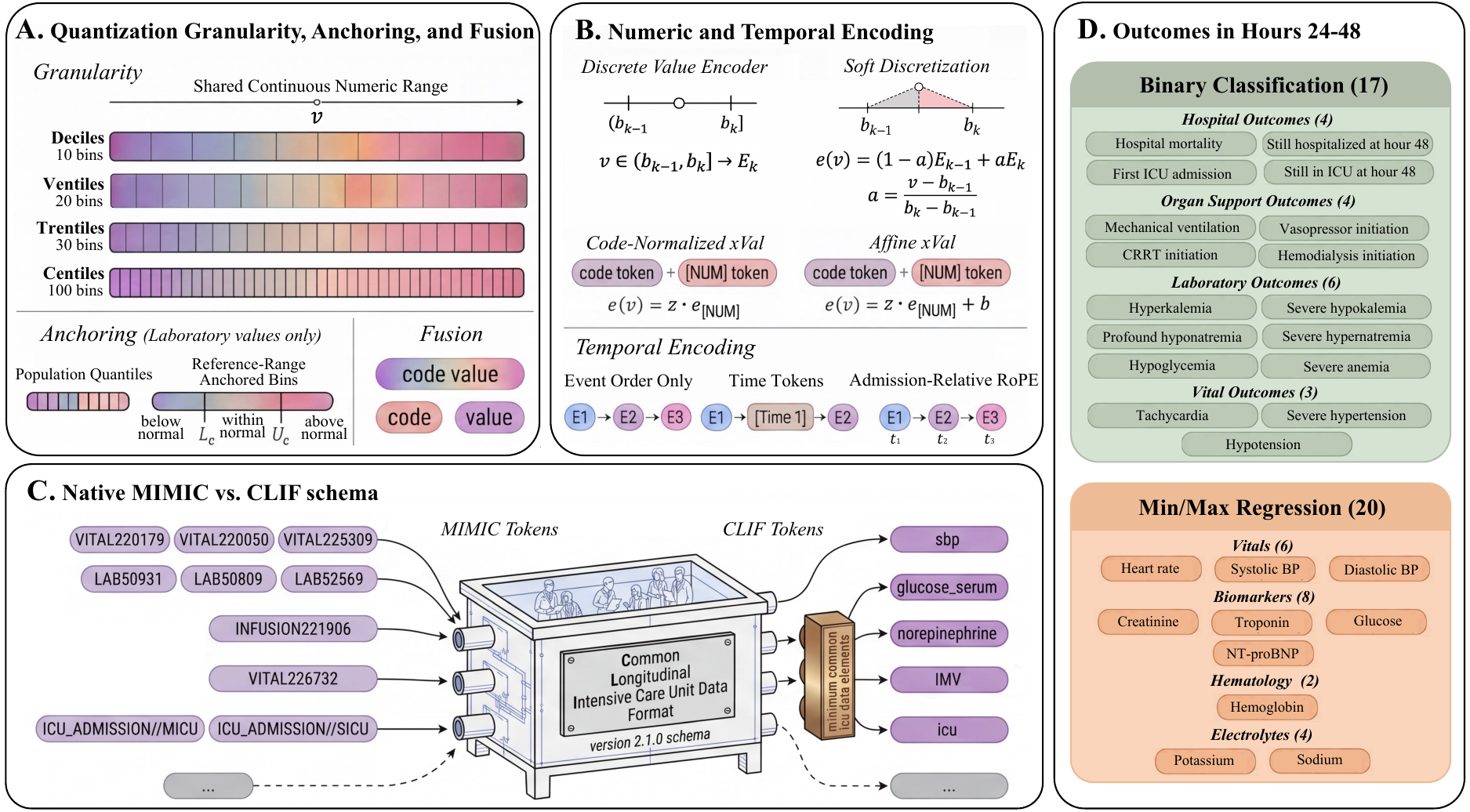}
		\caption{Panels A--C show Experiments~1--3 in order. Panel D shows the shared evaluation on outcomes in 24--48h.}
		\label{fig:main}
\end{figure*}

The core objective of this work is to identify useful starting configurations and to characterize how decisions made during tokenization affect downstream prediction under a shared training recipe. Concurrent work isolates three binary representation choices across pediatric tasks \citep{Guo26TokenizationTradeoffs}; here we add benchmarks of bin granularity, reference-range anchoring, continuous numeric encodings, temporal encodings, regression outcomes, and schema remapping into the Common Longitudinal Intensive Care Unit Data Format (CLIF)~2.1.0 \citep{Rojas25CLIF}. These experiments were performed across 156 training runs spanning both Llama-3.2-style and Qwen3-style architectures across three seeds on full MIMIC-IV hospitalizations. We evaluated frozen representations of the first 24 hours of hospitalization using linear probes on eight clinician-defined outcome families during hours 24--48 (\figureref{fig:main}). We examine these choices in three primary experiments:
\begin{enumerate}[nosep,leftmargin=*]
\item \textbf{Experiment~1 (Quantization and fusion)}: How should numeric values be quantized, and should code and value share a token? \textbf{Finding}: Fused code--value decile tokenization increased mean performance across all eight outcome families relative to unfused deciles, with statistically significant gains in six families at deciles (two binary and all four regression), and in laboratory AUROC at four other granularities. Finer quantization and reference-range anchoring did not yield consistent improvements, supporting fused deciles as a practical starting configuration.
\item \textbf{Experiment~2 (Numeric and temporal encoding)}: How should numeric magnitude and elapsed time be encoded? \textbf{Finding}: Soft discretization did not differ statistically from discrete deciles, whereas both code-normalized and affine xVal underperformed discrete deciles with statistically significant deficits across all eight families. Within discrete deciles, alternatives to explicit time tokens such as event order and admission-relative rotary position embeddings (RoPE) shortened sequences while marginally raising all eight family means (five and four statistically significant).
\item \textbf{Experiment~3 (Native MIMIC versus CLIF schema)}: How does remapping native MIMIC event streams to an expert-defined common data model change downstream prediction performance? \textbf{Finding}: The CLIF full-hospitalization training sequences contained 28.6\% as many tokens as the native sequences and 97.8\% fewer vocabulary entries, while producing statistically significant gains in four of eight families, no significant difference in two, and decreases in two.
\end{enumerate}

\section{Related Work}
\label{sec:related}

\subsection{Generative Medical Event Models and Their Benchmarks}
Early recurrent models such as Doctor AI, RETAIN, and Dipole predicted clinical outcomes from coded visit histories without laboratory or vital-sign values \citep{Choi16DoctorAI,Choi16RETAIN,Ma17Dipole}, and masked pretraining in BEHRT and Med-BERT kept the visit-level setup \citep{Li20BEHRT,Rasmy21MedBERT}. Subsequent sequence models operate directly on timestamped event streams through causal decoders \citep{McD23,Kra24}, state-space layers in EHRMamba \citep{Fallahpour24EHRMamba}, time-to-event heads in MOTOR \citep{Ste24MOTOR}, or text-serialized language models \citep{Hegselmann25LLMEHR}. Within the Medical Event Data Standard (MEDS) \citep{MEDS24}, ETHOS \citep{Ren24}, ETHOS-ARES \citep{Renc25ETHOSARES}, Curiosity \citep{Wax25}, and \citet{Burkhart25FoundationModels} all use unfused code and decile tokens.

Downstream benchmarks also fix input tokenization. EHRSHOT uses CLMBR with laboratory deciles \citep{Ste21CLMBR,Wor23}, and fixed preprocessing pipelines underlie HiRID-ICU-Benchmark \citep{Yeche21HiRIDICU}, multimodal MIMIC-IV evaluations \citep{Yu25JBHIBenchmark,Dai25CLIMB}, and EHRStruct \citep{Yang25EHRStruct}. Only \citet{Guo26TokenizationTradeoffs} directly compare fused and unfused tokens, interval tokens, age-derived RoPE, and workflow grouping across 74 pediatric tasks, and they find a consistent fused-token advantage. We extend controlled tokenization comparisons to models trained directly on adult hospitalizations, including an intensive care unit (ICU) cohort, and add regression outcomes, finer quantization, continuous encodings, and schema remapping.

\subsection{Encoding Magnitude and Time}
Encoding a scalar inside a discrete-token transformer remains an open problem in natural language processing (NLP) \citep{Thawani21Numbers} and is consequential in EHRs, where equal-frequency quantiles can split or merge clinically meaningful ranges \citep{Wax25}. Whereas xVal scales a shared \texttt{[NUM]} embedding and adds an auxiliary regression head \citep{Gol24}, Convex Combination of Semantic Embeddings (ConSE) forms convex combinations of category embeddings \citep{Nor14}; we adopt its local two-point interpolation between adjacent quantile embeddings. Other architectures predict continuous vectors over token chunks \citep{Sha25} or process continuous clinical values without a shared quantile table \citep{Labach23DuETT,Ma24TemporalCrossAttention}.

Timing information also varies across EHR models: CEHR-BERT inserts interval tokens and age embeddings \citep{Pan21}, Time2Vec learns continuous periodic features \citep{Kazemi19Time2Vec}, and MOTOR keys RoPE to patient age \citep{Ste24MOTOR,Su24RoFormer}. Across four MIMIC-IV tasks, however, \citet{Attrach25EHRTokenization} found no statistically significant change from dropping explicit time terms or substituting Time2Vec or Learnable Transformation-based Generalized Time Encoding (LeTE) \citep{Chen25LeTE}.

\subsection{Schema Standards}
Methods for medical code standardization act either on the clinical schema before tokenization or on the model embedding layer. Common data models such as the Observational Medical Outcomes Partnership (OMOP) \citep{Callahan23OMOP} and CLIF \citep{Rojas25CLIF} select clinical events, consolidate site-specific codes, and standardize units before tokenization, whereas code-embedding methods keep the source stream intact and change how each code is embedded. Some encode code text descriptions with a language model \citep{Hur22CHILCode} or attach knowledge-graph vectors \citep{Su25MedTok}. Others split International Classification of Diseases, Tenth Revision (ICD-10) and Systematized Nomenclature of Medicine--Clinical Terms (SNOMED CT) codes into subwords \citep{Dwivedi24MedStructRep}.

Our study evaluates end-to-end schema remapping from native MIMIC MEDS streams to CLIF~2.1.0 \citep{Rojas25CLIF,Liao26MIMICIVExtCLIF}. Remapping changes event selection, code consolidation, and token-frequency distributions at the same time (\sectionref{sec:exp3}).


\section{Methods}
\label{sec:methods}

\paragraph{Data and evaluation window.}
We convert each hospitalization (\texttt{hadm\_id}) into a chronological MEDS sequence~\citep{MEDS24} whose records store \texttt{subject\_id}, \texttt{time}, \texttt{code}, and, where applicable, \texttt{numeric\_value}. Decoders train on full hospitalizations, after which linear probes take frozen 24-hour representations to predict $(24,48]$ outcomes for stays still eligible at hour~24, using patient-level training, validation, and test splits (\appendixref{app:cohort}).

Experiments~1--2 evaluate 17 binary outcomes spanning hospital status, organ support, and physiologic abnormalities, including death, invasive mechanical ventilation initiation, and hyperkalemia. Twenty regression outcomes give the minimum and maximum follow-up values of vital signs and laboratory tests, such as heart rate, potassium, and creatinine (\tableref{tab:outcomes_sources}, \appendixref{app:appendix_results_tables}). Death during follow-up and hospital and ICU occupancy at hour 48 use all eligible hospitalizations. For first ICU admission, organ support, and physiologic thresholds, we exclude stays that already met the endpoint during the first 24 hours; the first two treat unrecorded follow-up events as negative, whereas threshold and regression tasks require an observed measurement in $(24,48]$.

Experiment~3 evaluates hospitalizations involving ICU care at any time, using a separate patient-level split (\appendixref{app:cohort}). Native MIMIC and CLIF inputs use identical hospitalization identifiers and labels derived from native MIMIC. Experiment~3 omits first ICU admission because every cohort hospitalization has an ICU stay, and keeps ICU occupancy at hour 48 (16 binary and 20 regression tasks, \tableref{tab:outcomes_sources}). \figureref{fig:main} summarizes the experiments.

\subsection{Representation Choices}

\paragraph{Quantization and fusion.} Numeric event streams listed in \appendixref{app:numeric_streams} are considered for numeric quantization. We partition each numeric code's values into 10, 20, 30, or 100 equal-frequency bins. Unfused representations use a separate code token followed by the numeric bin token. Fused tokens use one token per code--bin pair, increasing embedding parameters and vocabulary size. For laboratory measurements, reference-anchored alternatives allocate 5--10--5 or 10--10--10 bins below, within, and above the reference interval, with equal-frequency binning within each region. Reference limits come from MIMIC-IV and differ from the severe abnormality thresholds used as prediction targets. Experiment~1 crosses these six quantization schemes with both layouts, yielding 12 configurations.

\paragraph{Numeric encoding.} Experiment~2 compares discrete deciles, soft discretization, code-normalized xVal, and affine xVal. Where discrete deciles map each value to one quantile-bin token, soft discretization interpolates linearly between adjacent quantile embeddings from the value's position between their boundaries, assigning values below the first or above the last interior quantile cutoff to the first or last bin embedding \citep{Nor14}.

We evaluate two code-normalized implementations of xVal, which scale a shared numeric embedding~\citep{Gol24}. Each numeric event uses a code token followed by a numeric placeholder whose scalar subtracts the code-specific training median, divides by $\operatorname{IQR}/1.35$ (for interquartile range $\operatorname{IQR}$, minimum $10^{-8}$), and clips to $[-5, 5]$. Code-normalized xVal multiplies the placeholder vector by this normalized value, whereas affine xVal further adds a learned bias vector after scaling. Both include an auxiliary numeric-prediction head (\appendixref{app:value_encoders_formal}).

\paragraph{Temporal encoding.} We compare explicit time tokens, event order with sequential position indices, and admission-relative RoPE with elapsed time in 30-second steps \citep{Su24RoFormer}. Event order and admission-relative RoPE omit time tokens. Experiment~2 thus has 12 configurations, one for each pairing of the three temporal encodings with the four numeric encodings.

\subsection{Native MIMIC versus CLIF Schema}
\label{sec:exp3_methods}
Experiment~3 compares native MIMIC MEDS streams with their CLIF~2.1.0 conversion \citep{Rojas25CLIF,Liao26MIMICIVExtCLIF}. Each pipeline uses unfused deciles with explicit time tokens. CLIF inputs include admission, discharge, and transfer events, laboratory tests, vital signs, assessments, position, respiratory support, medications, continuous renal replacement therapy (CRRT), and code status. Hospitalizations and outcome labels are identical across pipelines (derived from native MIMIC). Event selection, code consolidation, and preprocessing vary between the pipelines, so this experiment evaluates the combined effect.

\begin{table*}[t]
\floatconts
  {tab:exp1_quantization}
  {\caption{Experiment~1: Unfused quantization relative to equal-frequency deciles. Across Tables~\ref{tab:exp1_quantization}--\ref{tab:exp3_family}, entries report six-model family means and signed differences ($\Delta$) with unadjusted 95\% confidence intervals ($n=6$, $\mathrm{df}=5$) in parentheses. Bold and {\boldmath$^{*}$} show BH-adjusted $p<0.05$.}}
  {\footnotesize
  \renewcommand{\arraystretch}{1.0}
  \setlength{\tabcolsep}{4.0pt}
  \begin{tabular}{@{}lcccccc@{}}
    \toprule
    & Reference & \multicolumn{2}{c}{Ventiles} & \multicolumn{2}{c}{Trentiles} & Centiles \\
    \cmidrule(lr){2-2} \cmidrule(lr){3-4} \cmidrule(lr){5-6} \cmidrule(lr){7-7}
    Family & Mean & $\Delta$ equal-frequency & $\Delta$ 5--10--5 & $\Delta$ equal-frequency & $\Delta$ 10--10--10 & $\Delta$ equal-frequency \\
    \midrule
    \multicolumn{7}{@{}l}{\textit{Binary families, AUROC}} \\
    \addlinespace[1pt]
    Hospital & $0.884$ & $+0.001$ & $+0.001$ & $+0.002$ & $0.000$ & $0.000$ \\[-2pt]
    & & {\scriptsize$(-\mkern-1.5mu 0.001{,}{+0.004})$} & {\scriptsize$(0.000{,}{+0.003})$} & {\scriptsize$(-\mkern-1.5mu 0.001{,}{+0.005})$} & {\scriptsize$(-\mkern-1.5mu 0.003{,}{+0.003})$} & {\scriptsize$(-\mkern-1.5mu 0.003{,}{+0.003})$} \\
    \addlinespace[1pt]
    Organ support & $0.958$ & $0.000$ & $0.000$ & $+0.001$ & $+0.001$ & $0.000$ \\[-2pt]
    & & {\scriptsize$(-\mkern-1.5mu 0.004{,}{+0.003})$} & {\scriptsize$(-\mkern-1.5mu 0.003{,}{+0.003})$} & {\scriptsize$(-\mkern-1.5mu 0.002{,}{+0.004})$} & {\scriptsize$(-\mkern-1.5mu 0.002{,}{+0.003})$} & {\scriptsize$(-\mkern-1.5mu 0.003{,}{+0.003})$} \\
    \addlinespace[1pt]
    Laboratory & $0.780$ & $-0.006$ & $-0.009$ & $+0.002$ & $-0.003$ & $-0.011$ \\[-2pt]
    & & {\scriptsize$(-\mkern-1.5mu 0.018{,}{+0.007})$} & {\scriptsize$(-\mkern-1.5mu 0.015{,}{-\mkern-1.5mu 0.003})$} & {\scriptsize$(-\mkern-1.5mu 0.009{,}{+0.013})$} & {\scriptsize$(-\mkern-1.5mu 0.010{,}{+0.005})$} & {\scriptsize$(-\mkern-1.5mu 0.019{,}{-\mkern-1.5mu 0.003})$} \\
    \addlinespace[1pt]
    Vital signs & $0.694$ & $+0.001$ & $-0.001$ & $+0.003$ & $+0.001$ & $0.000$ \\[-2pt]
    & & {\scriptsize$(-\mkern-1.5mu 0.001{,}{+0.004})$} & {\scriptsize$(-\mkern-1.5mu 0.010{,}{+0.008})$} & {\scriptsize$(-\mkern-1.5mu 0.004{,}{+0.009})$} & {\scriptsize$(-\mkern-1.5mu 0.004{,}{+0.006})$} & {\scriptsize$(-\mkern-1.5mu 0.004{,}{+0.004})$} \\
    \midrule
    \multicolumn{7}{@{}l}{\textit{Regression families, Spearman $\rho$}} \\
    \addlinespace[1pt]
    Electrolytes & $0.450$ & $-0.007$ & $-0.006$ & $-0.010$ & $-0.011$ & $-0.010$ \\[-2pt]
    & & {\scriptsize$(-\mkern-1.5mu 0.021{,}{+0.007})$} & {\scriptsize$(-\mkern-1.5mu 0.023{,}{+0.012})$} & {\scriptsize$(-\mkern-1.5mu 0.031{,}{+0.011})$} & {\scriptsize$(-\mkern-1.5mu 0.026{,}{+0.005})$} & {\scriptsize$(-\mkern-1.5mu 0.032{,}{+0.013})$} \\
    \addlinespace[1pt]
    Biomarkers & $0.437$ & $0.000$ & $-0.005$ & $0.000$ & $+0.005$ & $-0.011$ \\[-2pt]
    & & {\scriptsize$(-\mkern-1.5mu 0.017{,}{+0.018})$} & {\scriptsize$(-\mkern-1.5mu 0.025{,}{+0.015})$} & {\scriptsize$(-\mkern-1.5mu 0.009{,}{+0.008})$} & {\scriptsize$(-\mkern-1.5mu 0.010{,}{+0.019})$} & {\scriptsize$(-\mkern-1.5mu 0.025{,}{+0.004})$} \\
    \addlinespace[1pt]
    Hematology & $0.705$ & $+0.002$ & $-0.001$ & $-0.004$ & $-0.004$ & $-0.010$ \\[-2pt]
    & & {\scriptsize$(-\mkern-1.5mu 0.009{,}{+0.014})$} & {\scriptsize$(-\mkern-1.5mu 0.009{,}{+0.008})$} & {\scriptsize$(-\mkern-1.5mu 0.015{,}{+0.008})$} & {\scriptsize$(-\mkern-1.5mu 0.013{,}{+0.005})$} & {\scriptsize$(-\mkern-1.5mu 0.019{,}{-\mkern-1.5mu 0.001})$} \\
    \addlinespace[1pt]
    Vital signs & $0.465$ & $0.000$ & $+0.001$ & $+0.002$ & $0.000$ & $-0.002$ \\[-2pt]
    & & {\scriptsize$(-\mkern-1.5mu 0.010{,}{+0.011})$} & {\scriptsize$(-\mkern-1.5mu 0.010{,}{+0.012})$} & {\scriptsize$(-\mkern-1.5mu 0.004{,}{+0.007})$} & {\scriptsize$(-\mkern-1.5mu 0.003{,}{+0.004})$} & {\scriptsize$(-\mkern-1.5mu 0.012{,}{+0.007})$} \\
    \bottomrule
  \end{tabular}}
\end{table*}

\subsection{Training and Evaluation}
\paragraph{Models and training.} We initialize all parameters randomly and train scaled-down Llama-3.2-style \citep{Grattafiori24Llama3} and Qwen3-style \citep{Yang25Qwen3} causal decoders. Each decoder has 8 layers, 8 attention heads, a hidden dimension of 1024, an intermediate dimension of 2048, and a 4096-token context. With unfused deciles, this configuration has approximately 123.6M parameters for the Llama-3.2-style decoder and 103.7M for the Qwen3-style decoder. The Qwen3-style decoder ties input and output embeddings, whereas our Llama-3.2-style decoder does not. We train three seeds for each configuration, giving 72, 72, and 12 runs for Experiments~1--3.

All runs use AdamW with learning rate $10^{-4}$, weight decay 0.01, and minibatch size 4 for at most five epochs \citep{Gol24,Wax25}. Validation runs every half-epoch in Experiments~2--3 and at 10 equal step intervals in Experiment~1 (\appendixref{app:compute_by_stage}). A run stops once three consecutive evaluations show no validation-loss improvement. Experiment~1 uses packed sequences, whereas Experiments~2--3 use hospitalization-level windows (\appendixref{app:length_distributions}).

Discrete models minimize next-token cross-entropy. Soft discretization trains on weighted targets for adjacent bins, and xVal adds a numeric regression loss at numeric-placeholder positions. Because sequence lengths and vocabularies change across tokenizers, total token exposure, optimizer steps, and parameter counts also vary under this shared schedule (\tableref{tab:exp1_vocab_params}, \appendixref{app:compute_by_stage}).

\begin{table*}[t]
\floatconts
  {tab:exp1_fusion}
  {\caption{Experiment~1: Effect of code--value fusion ($\Delta$, fused minus unfused) at matched quantization granularity.}}
  {\footnotesize
  \renewcommand{\arraystretch}{1.0}
  \setlength{\tabcolsep}{4.0pt}
  \begin{tabular}{@{}lcccccc@{}}
    \toprule
    Family & Deciles & Ventiles & 5--10--5 & Trentiles & 10--10--10 & Centiles \\
    \midrule
    \multicolumn{7}{@{}l}{\textit{Binary families, AUROC}} \\
    \addlinespace[1pt]
    Hospital & {\boldmath$+0.004\rlap{$^{*}$}$} & {\boldmath$+0.003\rlap{$^{*}$}$} & {\boldmath$+0.004\rlap{$^{*}$}$} & $0.000$ & {\boldmath$+0.004\rlap{$^{*}$}$} & $+0.002$ \\[-2pt]
    & {\scriptsize\boldmath$(+0.001{,}{+0.007})$} & {\scriptsize\boldmath$(+0.001{,}{+0.004})$} & {\scriptsize\boldmath$(+0.002{,}{+0.005})$} & {\scriptsize$(-\mkern-1.5mu 0.001{,}{+0.002})$} & {\scriptsize\boldmath$(+0.002{,}{+0.005})$} & {\scriptsize$(-\mkern-1.5mu 0.002{,}{+0.005})$} \\
    \addlinespace[1pt]
    Organ support & $+0.002$ & $+0.001$ & $+0.001$ & $0.000$ & $+0.001$ & $+0.001$ \\[-2pt]
    & {\scriptsize$(0.000{,}{+0.004})$} & {\scriptsize$(-\mkern-1.5mu 0.005{,}{+0.006})$} & {\scriptsize$(-\mkern-1.5mu 0.002{,}{+0.003})$} & {\scriptsize$(-\mkern-1.5mu 0.002{,}{+0.002})$} & {\scriptsize$(-\mkern-1.5mu 0.002{,}{+0.003})$} & {\scriptsize$(-\mkern-1.5mu 0.002{,}{+0.004})$} \\
    \addlinespace[1pt]
    Laboratory & $+0.011$ & {\boldmath$+0.023\rlap{$^{*}$}$} & {\boldmath$+0.017\rlap{$^{*}$}$} & $+0.009$ & {\boldmath$+0.016\rlap{$^{*}$}$} & {\boldmath$+0.025\rlap{$^{*}$}$} \\[-2pt]
    & {\scriptsize$(0.000{,}{+0.022})$} & {\scriptsize\boldmath$(+0.009{,}{+0.036})$} & {\scriptsize\boldmath$(+0.011{,}{+0.023})$} & {\scriptsize$(-\mkern-1.5mu 0.009{,}{+0.028})$} & {\scriptsize\boldmath$(+0.004{,}{+0.027})$} & {\scriptsize\boldmath$(+0.015{,}{+0.036})$} \\
    \addlinespace[1pt]
    Vital signs & {\boldmath$+0.033\rlap{$^{*}$}$} & {\boldmath$+0.032\rlap{$^{*}$}$} & {\boldmath$+0.031\rlap{$^{*}$}$} & {\boldmath$+0.028\rlap{$^{*}$}$} & {\boldmath$+0.026\rlap{$^{*}$}$} & {\boldmath$+0.027\rlap{$^{*}$}$} \\[-2pt]
    & {\scriptsize\boldmath$(+0.026{,}{+0.039})$} & {\scriptsize\boldmath$(+0.028{,}{+0.036})$} & {\scriptsize\boldmath$(+0.023{,}{+0.039})$} & {\scriptsize\boldmath$(+0.024{,}{+0.031})$} & {\scriptsize\boldmath$(+0.022{,}{+0.031})$} & {\scriptsize\boldmath$(+0.023{,}{+0.032})$} \\
    \midrule
    \multicolumn{7}{@{}l}{\textit{Regression families, Spearman $\rho$}} \\
    \addlinespace[1pt]
    Electrolytes & {\boldmath$+0.069\rlap{$^{*}$}$} & {\boldmath$+0.072\rlap{$^{*}$}$} & {\boldmath$+0.062\rlap{$^{*}$}$} & {\boldmath$+0.066\rlap{$^{*}$}$} & {\boldmath$+0.069\rlap{$^{*}$}$} & {\boldmath$+0.063\rlap{$^{*}$}$} \\[-2pt]
    & {\scriptsize\boldmath$(+0.050{,}{+0.088})$} & {\scriptsize\boldmath$(+0.057{,}{+0.088})$} & {\scriptsize\boldmath$(+0.041{,}{+0.083})$} & {\scriptsize\boldmath$(+0.050{,}{+0.083})$} & {\scriptsize\boldmath$(+0.063{,}{+0.075})$} & {\scriptsize\boldmath$(+0.044{,}{+0.082})$} \\
    \addlinespace[1pt]
    Biomarkers & {\boldmath$+0.032\rlap{$^{*}$}$} & {\boldmath$+0.024\rlap{$^{*}$}$} & {\boldmath$+0.031\rlap{$^{*}$}$} & {\boldmath$+0.028\rlap{$^{*}$}$} & {\boldmath$+0.011\rlap{$^{*}$}$} & {\boldmath$+0.020\rlap{$^{*}$}$} \\[-2pt]
    & {\scriptsize\boldmath$(+0.015{,}{+0.048})$} & {\scriptsize\boldmath$(+0.016{,}{+0.032})$} & {\scriptsize\boldmath$(+0.020{,}{+0.041})$} & {\scriptsize\boldmath$(+0.025{,}{+0.032})$} & {\scriptsize\boldmath$(+0.001{,}{+0.021})$} & {\scriptsize\boldmath$(+0.011{,}{+0.030})$} \\
    \addlinespace[1pt]
    Hematology & {\boldmath$+0.099\rlap{$^{*}$}$} & {\boldmath$+0.090\rlap{$^{*}$}$} & {\boldmath$+0.099\rlap{$^{*}$}$} & {\boldmath$+0.093\rlap{$^{*}$}$} & {\boldmath$+0.088\rlap{$^{*}$}$} & {\boldmath$+0.091\rlap{$^{*}$}$} \\[-2pt]
    & {\scriptsize\boldmath$(+0.091{,}{+0.107})$} & {\scriptsize\boldmath$(+0.078{,}{+0.101})$} & {\scriptsize\boldmath$(+0.089{,}{+0.110})$} & {\scriptsize\boldmath$(+0.082{,}{+0.104})$} & {\scriptsize\boldmath$(+0.072{,}{+0.105})$} & {\scriptsize\boldmath$(+0.075{,}{+0.106})$} \\
    \addlinespace[1pt]
    Vital signs & {\boldmath$+0.114\rlap{$^{*}$}$} & {\boldmath$+0.111\rlap{$^{*}$}$} & {\boldmath$+0.111\rlap{$^{*}$}$} & {\boldmath$+0.106\rlap{$^{*}$}$} & {\boldmath$+0.108\rlap{$^{*}$}$} & {\boldmath$+0.107\rlap{$^{*}$}$} \\[-2pt]
    & {\scriptsize\boldmath$(+0.109{,}{+0.118})$} & {\scriptsize\boldmath$(+0.103{,}{+0.119})$} & {\scriptsize\boldmath$(+0.103{,}{+0.119})$} & {\scriptsize\boldmath$(+0.102{,}{+0.111})$} & {\scriptsize\boldmath$(+0.102{,}{+0.113})$} & {\scriptsize\boldmath$(+0.099{,}{+0.116})$} \\
    \bottomrule
  \end{tabular}}
\end{table*}

\paragraph{Linear probes and evaluation metrics.} We extract the final-layer hidden state at the last token of the 24-hour input before any end-of-sequence, context-truncation, or padding tokens. For each outcome, we standardize the input features with the mean and standard deviation of that outcome's eligible training hospitalizations. Linear probes use logistic regression ($C=0.01$) for binary outcomes and Ridge regression ($\alpha=100$) for regression outcomes. For Ridge probes, we also standardize each target with the mean and standard deviation from those eligible training hospitalizations. We then rescale the predicted values to their original clinical units for evaluation.

Binary outcomes are scored with area under the receiver operating characteristic curve (AUROC), Brier score, and 15-bin expected calibration error. Regression outcomes are scored with Spearman correlation ($\rho$), coefficient of determination ($R^2$), mean absolute error, and root mean squared error (\appendixref{app:cohort}). Main tables report AUROC or Spearman correlation ($\rho$) depending on outcome type, spanning four clinician-defined binary families (hospital, organ support, laboratory, and vital signs) and four regression families (electrolytes, biomarkers, hematology, and vital signs). For each model, we average metrics across outcomes within the family and then average across the two architectures and three initialization seeds (\appendixref{app:result_tables}).

\paragraph{Statistical comparisons.}
\label{sec:stats_compare}
We compare family-mean performance by pairing each model with its baseline on architecture and seed ($n=6$ pairs) in a two-sided one-sample $t$-test ($\mathrm{df}=5$) with unadjusted 95\% $t$-intervals. In Tables~\ref{tab:exp1_quantization}--\ref{tab:exp3_family}, 17 $\Delta$ columns over the eight outcome families give 136 comparisons; we apply Benjamini--Hochberg (BH) adjustment at $q<0.05$ separately within quantization (40 tests), fusion (48), numeric encoding (24), temporal encoding (16), and schema remapping (8), marking BH $p<0.05$ in bold with {\boldmath$^{*}$} \citep{Benjamini95FDR}. Two-sided Welch $t$-tests under the same BH adjustment compare Llama-3.2-style and Qwen3-style differences, and a 1{,}000-draw patient-clustered bootstrap checks test-set sampling uncertainty given that 75.1\% of test hospitalizations in Experiments~1--2 come from repeat-visit patients (\appendixref{app:statistical-comparisons}).

\section{Results}
\label{sec:results}
Across the 136 family comparisons in Tables~\ref{tab:exp1_quantization}--\ref{tab:exp3_family}, 74 unadjusted 95\% $t$-intervals exclude zero ($p<0.05$) and 69 remain significant after BH adjustment ($p<0.05$), all 69 of which also exclude zero under the patient-clustered bootstrap (\appendixref{app:statistical-comparisons}). None of the 136 family comparisons differed significantly between Llama-3.2-style and Qwen3-style decoders (Welch $t$-test, BH $p\geq 0.56$; full tables in \appendixref{app:result_tables}).

\begin{table*}[t]
\floatconts
  {tab:exp2_numeric}
  {\caption{Experiment~2: Numeric and temporal encodings ($\Delta$ relative to discrete deciles with explicit time tokens). Numeric encoding columns vary the value representation with explicit time tokens fixed. Temporal encoding columns vary the time representation with discrete deciles fixed.}}
  {\footnotesize
  \renewcommand{\arraystretch}{1.0}
  \setlength{\tabcolsep}{4.0pt}
  \begin{tabular}{@{}lcccccc@{}}
    \toprule
    & Reference & \multicolumn{3}{c}{Numeric encoding $\Delta$} & \multicolumn{2}{c}{Temporal encoding $\Delta$} \\
    \cmidrule(lr){2-2} \cmidrule(lr){3-5} \cmidrule(lr){6-7}
    & & Soft & Code-normalized & Affine & Event & Admission-relative \\[-1pt]
    Family & Mean & discretization & xVal & xVal & order & RoPE \\
    \midrule
    \multicolumn{7}{@{}l}{\textit{Binary families, AUROC}} \\
    \addlinespace[1pt]
    Hospital & $0.890$ & $-0.001$ & {\boldmath$-0.050\rlap{$^{*}$}$} & {\boldmath$-0.052\rlap{$^{*}$}$} & {\boldmath$+0.005\rlap{$^{*}$}$} & {\boldmath$+0.005\rlap{$^{*}$}$} \\[-2pt]
    & & {\scriptsize$(-\mkern-1.5mu 0.004{,}{+0.003})$} & {\scriptsize\boldmath$(-\mkern-1.5mu 0.054{,}{-\mkern-1.5mu 0.047})$} & {\scriptsize\boldmath$(-\mkern-1.5mu 0.054{,}{-\mkern-1.5mu 0.049})$} & {\scriptsize\boldmath$(+0.003{,}{+0.007})$} & {\scriptsize\boldmath$(+0.001{,}{+0.008})$} \\
    \addlinespace[1pt]
    Organ support & $0.959$ & $-0.001$ & {\boldmath$-0.057\rlap{$^{*}$}$} & {\boldmath$-0.057\rlap{$^{*}$}$} & {\boldmath$+0.003\rlap{$^{*}$}$} & {\boldmath$+0.002\rlap{$^{*}$}$} \\[-2pt]
    & & {\scriptsize$(-\mkern-1.5mu 0.002{,}{+0.001})$} & {\scriptsize\boldmath$(-\mkern-1.5mu 0.060{,}{-\mkern-1.5mu 0.055})$} & {\scriptsize\boldmath$(-\mkern-1.5mu 0.059{,}{-\mkern-1.5mu 0.055})$} & {\scriptsize\boldmath$(+0.002{,}{+0.005})$} & {\scriptsize\boldmath$(0.000{,}{+0.003})$} \\
    \addlinespace[1pt]
    Laboratory & $0.785$ & $-0.002$ & {\boldmath$-0.056\rlap{$^{*}$}$} & {\boldmath$-0.051\rlap{$^{*}$}$} & $+0.009$ & $+0.008$ \\[-2pt]
    & & {\scriptsize$(-\mkern-1.5mu 0.021{,}{+0.017})$} & {\scriptsize\boldmath$(-\mkern-1.5mu 0.062{,}{-\mkern-1.5mu 0.050})$} & {\scriptsize\boldmath$(-\mkern-1.5mu 0.060{,}{-\mkern-1.5mu 0.043})$} & {\scriptsize$(-\mkern-1.5mu 0.010{,}{+0.028})$} & {\scriptsize$(+0.001{,}{+0.015})$} \\
    \addlinespace[1pt]
    Vital signs & $0.701$ & $0.000$ & {\boldmath$-0.054\rlap{$^{*}$}$} & {\boldmath$-0.053\rlap{$^{*}$}$} & {\boldmath$+0.006\rlap{$^{*}$}$} & $+0.007$ \\[-2pt]
    & & {\scriptsize$(-\mkern-1.5mu 0.005{,}{+0.005})$} & {\scriptsize\boldmath$(-\mkern-1.5mu 0.057{,}{-\mkern-1.5mu 0.050})$} & {\scriptsize\boldmath$(-\mkern-1.5mu 0.059{,}{-\mkern-1.5mu 0.047})$} & {\scriptsize\boldmath$(+0.001{,}{+0.011})$} & {\scriptsize$(+0.001{,}{+0.013})$} \\
    \midrule
    \multicolumn{7}{@{}l}{\textit{Regression families, Spearman $\rho$}} \\
    \addlinespace[1pt]
    Electrolytes & $0.471$ & $-0.011$ & {\boldmath$-0.125\rlap{$^{*}$}$} & {\boldmath$-0.112\rlap{$^{*}$}$} & $+0.019$ & $+0.019$ \\[-2pt]
    & & {\scriptsize$(-\mkern-1.5mu 0.032{,}{+0.009})$} & {\scriptsize\boldmath$(-\mkern-1.5mu 0.153{,}{-\mkern-1.5mu 0.097})$} & {\scriptsize\boldmath$(-\mkern-1.5mu 0.129{,}{-\mkern-1.5mu 0.094})$} & {\scriptsize$(-\mkern-1.5mu 0.002{,}{+0.040})$} & {\scriptsize$(-\mkern-1.5mu 0.004{,}{+0.042})$} \\
    \addlinespace[1pt]
    Biomarkers & $0.458$ & $+0.003$ & {\boldmath$-0.065\rlap{$^{*}$}$} & {\boldmath$-0.066\rlap{$^{*}$}$} & $+0.010$ & $+0.006$ \\[-2pt]
    & & {\scriptsize$(-\mkern-1.5mu 0.010{,}{+0.015})$} & {\scriptsize\boldmath$(-\mkern-1.5mu 0.086{,}{-\mkern-1.5mu 0.044})$} & {\scriptsize\boldmath$(-\mkern-1.5mu 0.081{,}{-\mkern-1.5mu 0.051})$} & {\scriptsize$(0.000{,}{+0.021})$} & {\scriptsize$(-\mkern-1.5mu 0.010{,}{+0.023})$} \\
    \addlinespace[1pt]
    Hematology & $0.731$ & $-0.007$ & {\boldmath$-0.162\rlap{$^{*}$}$} & {\boldmath$-0.157\rlap{$^{*}$}$} & {\boldmath$+0.014\rlap{$^{*}$}$} & {\boldmath$+0.019\rlap{$^{*}$}$} \\[-2pt]
    & & {\scriptsize$(-\mkern-1.5mu 0.018{,}{+0.004})$} & {\scriptsize\boldmath$(-\mkern-1.5mu 0.167{,}{-\mkern-1.5mu 0.158})$} & {\scriptsize\boldmath$(-\mkern-1.5mu 0.161{,}{-\mkern-1.5mu 0.153})$} & {\scriptsize\boldmath$(+0.008{,}{+0.020})$} & {\scriptsize\boldmath$(+0.012{,}{+0.025})$} \\
    \addlinespace[1pt]
    Vital signs & $0.473$ & $-0.004$ & {\boldmath$-0.094\rlap{$^{*}$}$} & {\boldmath$-0.096\rlap{$^{*}$}$} & {\boldmath$+0.014\rlap{$^{*}$}$} & {\boldmath$+0.018\rlap{$^{*}$}$} \\[-2pt]
    & & {\scriptsize$(-\mkern-1.5mu 0.016{,}{+0.007})$} & {\scriptsize\boldmath$(-\mkern-1.5mu 0.105{,}{-\mkern-1.5mu 0.083})$} & {\scriptsize\boldmath$(-\mkern-1.5mu 0.108{,}{-\mkern-1.5mu 0.084})$} & {\scriptsize\boldmath$(+0.007{,}{+0.020})$} & {\scriptsize\boldmath$(+0.008{,}{+0.029})$} \\
    \bottomrule
  \end{tabular}}
\end{table*}

\subsection{Experiment 1: Quantization and Code--Value Fusion}
\label{sec:exp1_results}

Fused tokenization produced the largest and most consistent family-mean differences (\tableref{tab:exp1_fusion}, \tableref{tab:appendix_family_means_exp1}). At matched deciles, fused and unfused tokenizers share bin boundaries: unfused events use adjacent code and shared bin tokens, whereas fused events use one code-specific value token. Fusion raised the six-model point estimate for all eight families ($+0.002$ to $+0.033$ AUROC, $+0.032$ to $+0.114$ Spearman $\rho$), with statistically significant gains in six families: hospital AUROC ($+0.004$, $t=3.61$, BH $p=0.020$), vital-signs AUROC ($+0.033$, $t=12.98$, BH $p=1.2\times 10^{-4}$), and all four regression families ($t=4.79$ to $67.76$, BH $p\leq 0.0072$). Organ-support AUROC ($+0.002$, $t=2.28$, BH $p=0.086$) and laboratory AUROC ($+0.011$, $t=2.54$, BH $p=0.064$) were not statistically significant at deciles, although laboratory AUROC was significant at four other granularities ($+0.016$ to $+0.025$, BH $p\leq 0.022$). Across all six granularities, 38 of 48 fusion comparisons were statistically significant, which covers seven of eight families (organ-support AUROC at none).

At the outcome level (\tableref{tab:appendix_outcome_means_exp1}), decile fusion raised 20 of 20 regression means (all 20 BH-significant) and 16 of 17 binary means (eight BH-significant). The largest regression gains came from minimum and maximum heart rate ($+0.249$ and $+0.194$, from $0.445$ to $0.694$ and $0.451$ to $0.644$), sodium ($+0.103$ and $+0.107$), and hemoglobin ($+0.096$ and $+0.102$). Binary gains were largest on tachycardia ($+0.046$), severe anemia ($+0.042$), severe hypertension ($+0.040$), and hypoglycemia ($+0.027$), whereas hyperkalemia was the lone decrease ($-0.023$, BH $p=0.042$).

Finer unfused quantiles and reference-range anchoring gave small, non-monotonic differences from unfused deciles (\tableref{tab:exp1_quantization}). Three of 40 unadjusted 95\% intervals excluded zero ($p=0.014$ to $0.036$), all decreases, but none survived BH adjustment ($\Delta$ from $-0.011$ to $+0.003$ AUROC and $-0.011$ to $+0.005$ Spearman $\rho$, $|t|\leq 3.71$, BH $p\geq 0.33$).

Even without numeric distance supervision under cross-entropy training, centile token embeddings still organized along numeric and clinical scales (\appendixref{app:representation_analyses}): principal-component projections trace smooth ordered curves from the lowest to the highest centile in both unfused and fused models (\figureref{fig:pca_centile_geometry}), and linear probes on unfused centile embeddings classify bins as within or outside MIMIC-IV reference ranges at 93--97\% accuracy without reference-interval supervision (\figureref{fig:clinical_boundary_probe}). On decile and ventile embeddings, the same linear probes never exceed majority-class accuracy. But neither learned centile ordinality nor linearly decodable reference ranges in token space improved downstream 24-hour probes.

\subsection{Experiment 2: Numeric and Temporal Encoding}
\label{sec:exp2}

\paragraph{Numeric encoding.} Experiment~2 crosses four numeric encodings with three temporal encodings at unfused population deciles (\tableref{tab:appendix_family_means_exp2}, \tableref{tab:appendix_outcome_means_exp2}). With explicit time tokens fixed, soft discretization showed no statistically significant difference from discrete deciles in any family ($\Delta$ from $-0.002$ to $0.000$ AUROC and $-0.011$ to $+0.003$ Spearman $\rho$, $|t|\leq 1.71$, BH $p\geq 0.21$, \tableref{tab:exp2_numeric}). Code-normalized xVal and affine xVal were lower than discrete deciles in all eight families (16 of 16 comparisons significant, BH $p\leq 7.6\times 10^{-4}$). Code-normalized xVal fell by $0.050$ to $0.057$ AUROC ($t=-22.91$ to $-52.62$) and $0.065$ to $0.162$ Spearman $\rho$ ($t=-7.96$ to $-87.94$). Affine xVal similarly fell by $0.051$ to $0.057$ AUROC and $0.066$ to $0.157$ Spearman $\rho$. Outcome-level regression decreases under xVal were largest on minimum and maximum hemoglobin ($-0.156$ to $-0.163$), sodium ($-0.114$ to $-0.129$), and potassium ($-0.109$ to $-0.122$).

\paragraph{Temporal encoding with discrete deciles.}
Holding discrete deciles fixed, both event order and admission-relative RoPE raised point estimates over explicit time tokens in all eight families, but only marginally (at most $+0.009$ AUROC and $+0.019$ Spearman $\rho$; \tableref{tab:exp2_numeric}). Event order gave statistically significant gains in five of eight families ($+0.003$ to $+0.006$ AUROC, $+0.014$ Spearman $\rho$, $t=3.21$ to $6.69$, BH $p\leq 0.047$). Admission-relative RoPE did so in four families ($+0.002$ to $+0.005$ AUROC, $+0.018$ to $+0.019$ Spearman $\rho$, $t=3.11$ to $7.01$, BH $p\leq 0.047$). Under event order and admission-relative RoPE, laboratory AUROC ($+0.009$ and $+0.008$, unadjusted $p=0.29$ and $0.038$, BH $p=0.31$ and $0.055$), electrolytes $\rho$ ($+0.019$ and $+0.019$, BH $p=0.083$ and $0.094$), and biomarkers $\rho$ ($+0.010$ and $+0.006$, BH $p=0.077$ and $0.38$) were not significant, nor was vital-signs AUROC under admission-relative RoPE ($+0.007$, $t=2.97$, unadjusted $p=0.031$, BH $p=0.050$~[$>0.05$]). Omitting time tokens reduced median first-24h sequence length from 93 to 83 tokens, approximately 11\% (\figureref{fig:temporal_len_distribution}).

\subsection{Experiment~3: Native MIMIC versus CLIF Schema}
\label{sec:exp3}
Across the 57{,}690 training hospitalizations, the full-hospitalization training split contained 684.7 million native MIMIC tokens and 195.9 million CLIF tokens (28.6\% of the native total, or 71.4\% fewer tokens), and schema remapping reduced the tokenizer vocabulary from 17{,}752 to 397 entries (97.8\% fewer entries). Remapping to CLIF significantly improved four of the eight outcome families (\tableref{tab:exp3_family}, \tableref{tab:appendix_family_means_exp3}): hospital AUROC rose by $0.010$ ($t=9.48$, BH $p=4.4\times 10^{-4}$), vital-signs AUROC by $0.032$ ($t=11.38$, BH $p=2.5\times 10^{-4}$), hematology $\rho$ by $0.029$ ($t=4.51$, BH $p=0.0085$), and vital-signs $\rho$ by $0.099$ ($t=22.49$, BH $p=2.6\times 10^{-5}$). Laboratory AUROC ($+0.006$, $t=0.65$, BH $p=0.62$) and biomarkers $\rho$ ($+0.003$, $t=0.42$, BH $p=0.69$) were statistically unchanged. Organ-support AUROC fell by $0.009$ ($t=-7.05$, BH $p=0.0014$), and electrolytes $\rho$ fell by $0.035$ ($t=-11.33$, BH $p=2.5\times 10^{-4}$).

\begin{table}[t]
\floatconts
  {tab:exp3_family}
  {\caption{Experiment~3: Effect of schema remapping ($\Delta$, CLIF~2.1.0 minus native MIMIC).}}
  {\footnotesize
  \renewcommand{\arraystretch}{1.0}
  \setlength{\tabcolsep}{7.0pt}
  \begin{tabular}{@{}lcc@{}}
    \toprule
    & Reference & $\Delta$ \\
    \cmidrule(lr){2-2} \cmidrule(lr){3-3}
    Family & Native MIMIC & CLIF 2.1.0 \\
    \midrule
    \multicolumn{3}{@{}l}{\textit{Binary families, AUROC}} \\
    \addlinespace[1pt]
    Hospital & $0.877$ & {\boldmath$+0.010\rlap{$^{*}$}$} \\[-2pt]
    & & {\scriptsize\boldmath$(+0.007{,}{+0.012})$} \\
    \addlinespace[1pt]
    Organ support & $0.926$ & {\boldmath$-0.009\rlap{$^{*}$}$} \\[-2pt]
    & & {\scriptsize\boldmath$(-\mkern-1.5mu 0.012{,}{-\mkern-1.5mu 0.006})$} \\
    \addlinespace[1pt]
    Laboratory & $0.729$ & $+0.006$ \\[-2pt]
    & & {\scriptsize$(-\mkern-1.5mu 0.019{,}{+0.031})$} \\
    \addlinespace[1pt]
    Vital signs & $0.711$ & {\boldmath$+0.032\rlap{$^{*}$}$} \\[-2pt]
    & & {\scriptsize\boldmath$(+0.025{,}{+0.039})$} \\
    \midrule
    \multicolumn{3}{@{}l}{\textit{Regression families, Spearman $\rho$}} \\
    \addlinespace[1pt]
    Electrolytes & $0.425$ & {\boldmath$-0.035\rlap{$^{*}$}$} \\[-2pt]
    & & {\scriptsize\boldmath$(-\mkern-1.5mu 0.043{,}{-\mkern-1.5mu 0.027})$} \\
    \addlinespace[1pt]
    Biomarkers & $0.425$ & $+0.003$ \\[-2pt]
    & & {\scriptsize$(-\mkern-1.5mu 0.016{,}{+0.022})$} \\
    \addlinespace[1pt]
    Hematology & $0.635$ & {\boldmath$+0.029\rlap{$^{*}$}$} \\[-2pt]
    & & {\scriptsize\boldmath$(+0.012{,}{+0.045})$} \\
    \addlinespace[1pt]
    Vital signs & $0.494$ & {\boldmath$+0.099\rlap{$^{*}$}$} \\[-2pt]
    & & {\scriptsize\boldmath$(+0.088{,}{+0.110})$} \\
    \bottomrule
  \end{tabular}}
\end{table}

Among individual outcomes (\tableref{tab:appendix_outcome_means_exp3}), heart-rate extrema ($+0.157$ and $+0.120$) and diastolic and systolic blood pressure ($+0.057$ to $+0.091$) showed the largest improvements. The largest decreases came from hemodialysis initiation AUROC ($-0.050$), hypoglycemia AUROC ($-0.040$), and minimum potassium $\rho$ ($-0.075$).

\section{Discussion}
\label{sec:discussion}

Tokenization substantially affected downstream prediction performance under a shared training recipe across configurations. As a whole, the results support three practical takeaways: prioritize code--value fusion, do not assume continuous encodings will inherently improve performance, and consider whether expert-curated common data models help focus on high-signal clinical events.

\paragraph{Code--value fusion.}
Fused tokens improved mean performance across all eight outcome families, with statistically significant gains in six families at deciles and 38 of 48 comparisons across all six granularities (seven of eight families, \tableref{tab:exp1_fusion}). These gains substantially generalize the fused-token advantage reported by \citet{Guo26TokenizationTradeoffs} to models trained directly on adult hospitalizations and evaluated on both binary and regression outcomes. Finer quantization and reference-range anchoring produced no consistent benefit, supporting fused deciles as a practical starting configuration. Fusion changes vocabulary size, sequence length, and embedding capacity together (\tableref{tab:exp1_vocab_params}), so future work is needed to isolate attribution. That said, increasing embedding size is unlikely to be the source of the fusion gain: fused centiles further added 109.4M and 54.7M parameters over fused deciles on Llama-3.2-style and Qwen3-style decoders, but they lowered seven family means ($-0.001$ to $-0.022$, \tableref{tab:appendix_family_means_exp1}). Context-window truncation is also unlikely to explain the fusion gain, as the 4096-token context already covered more than 99.95\% of first-24h sequences for both fused and unfused tokenizations (\figureref{fig:seq_len_distribution}).

\paragraph{Task-level sensitivity.}
Which individual outcomes moved depended on the representation choice under test, not on one fixed subset of tasks (\appendixref{app:result_tables}). On follow-up extrema for heart rate, sodium, and hemoglobin, fusion raised Spearman $\rho$ across all six granularities ($+0.188$ to $+0.260$, $+0.092$ to $+0.108$, and $+0.086$ to $+0.102$), whereas both xVal variants reduced those six correlations ($-0.105$ to $-0.163$) and unfused granularity or anchoring shifted them by $-0.012$ to $+0.006$. Event order and admission-relative RoPE added $+0.011$ to $+0.019$ on all six extrema; CLIF raised heart rate ($+0.157$, $+0.120$) and hemoglobin ($+0.027$, $+0.030$) but lowered sodium ($-0.009$, $-0.017$). Neither decoder architecture showed systematically larger shifts.

\paragraph{Numeric and temporal encoding complexity.} Soft discretization did not differ significantly from discrete deciles, while the two xVal variants fell below discrete deciles in all eight families. An affine bias did not close that gap, leaving per-code normalization, auxiliary loss weighting, and representation geometry as open factors. Event order and admission-relative RoPE shortened inputs and gave marginal but significant gains in five and four of eight families over explicit time tokens, so discrete deciles with either encoding form remain a practical single-stay baseline.

\paragraph{Expert-curated schemas and vocabulary compression.} Schema remapping to CLIF cut training tokens by 71.4\% and vocabulary entries by 97.8\%, yet still improved four outcome families (\tableref{tab:exp3_family}). Two factors help explain these gains. First, dozens of clinician-scientists defined CLIF to preserve high-signal critical-care concepts and standardize units \citep{Rojas25CLIF}. Second, a smaller vocabulary can help in clinical settings, which have far less data than many other transformer domains.

\paragraph{Limitations.}
Training 156 transformers is computationally expensive and limited the hyperparameter sweeps we could run, so all configurations shared one optimization schedule and model size. Our results also come from a single site (MIMIC-IV), so multi-site tests with shared schemas such as CLIF remain needed. Choices such as code--value fusion and schema remapping alter sequence length, vocabulary size, and parameter counts at the same time, so we estimate their combined effects. We also score frozen 24-hour representations with linear probes rather than generative rollouts.

\section{Conclusion}
\label{sec:conclusion}
Tokenization is a consequential design choice for generative medical event models. Under a shared training recipe, changing how we encode clinical events produced significant downstream differences in linear-probe performance. Fused code--value deciles consistently improved binary and regression probes over unfused deciles, whereas finer quantization, reference-range anchoring, and xVal gave no consistent advantage. Replacing time tokens with event order or admission-relative RoPE shortened sequences with small downstream gains, and CLIF remapping compressed the vocabulary and token stream with significant gains in four families, no significant change in two, and decreases in two.

Our results offer actionable configurations for model tokenization and encourage further testing to see whether the observed benefits generalize across institutions and generative rollouts. When building generative medical event models, tokenization should be assessed alongside architectural and training decisions. Because the tokenizer effectively determines the upper bound of what next-token learning can recover from an EHR sequence, evaluating it is among the most cost-effective steps to take before scaling either the model or the dataset.

\iffinal
\acks{Compute resources for this study were provided by the Randi HPC Cluster maintained by the Center for Research Informatics (CRI) at the University of Chicago. The Center for Research Informatics is funded by the Biological Sciences Division and the Institute for Translational Medicine/CTSA (NIH UL1TR002389) at the University of Chicago.}
\fi

\begingroup
\raggedright
\bibliography{references_non_arxiv,references}
\endgroup

\clearpage
\onecolumn
\appendix

\section{Representation Analyses}
\label{app:representation_analyses}

\subsection{Embedding Geometry: Shared and Code-Specific Centile Embeddings}
\label{app:embedding_geometry}

\begin{figure*}[htbp]
\floatconts
  {fig:pca_centile_geometry}
  {\caption{Embedding geometry from the Llama-3.2-style decoder. Left: principal-component projection of the shared unfused centile embeddings. Right: fused, code-specific centile embeddings for six measurements. Colors and dashed segments follow centile order from the lowest bin (\texttt{Q0}) to the highest (\texttt{Q99}). Each panel uses the checkpoint selected by validation loss. Panel labels give the count of distinct bins (realized bins), which is less than 100 when rounded measurements share quantile boundaries.}}
  {\includegraphics[width=\linewidth]{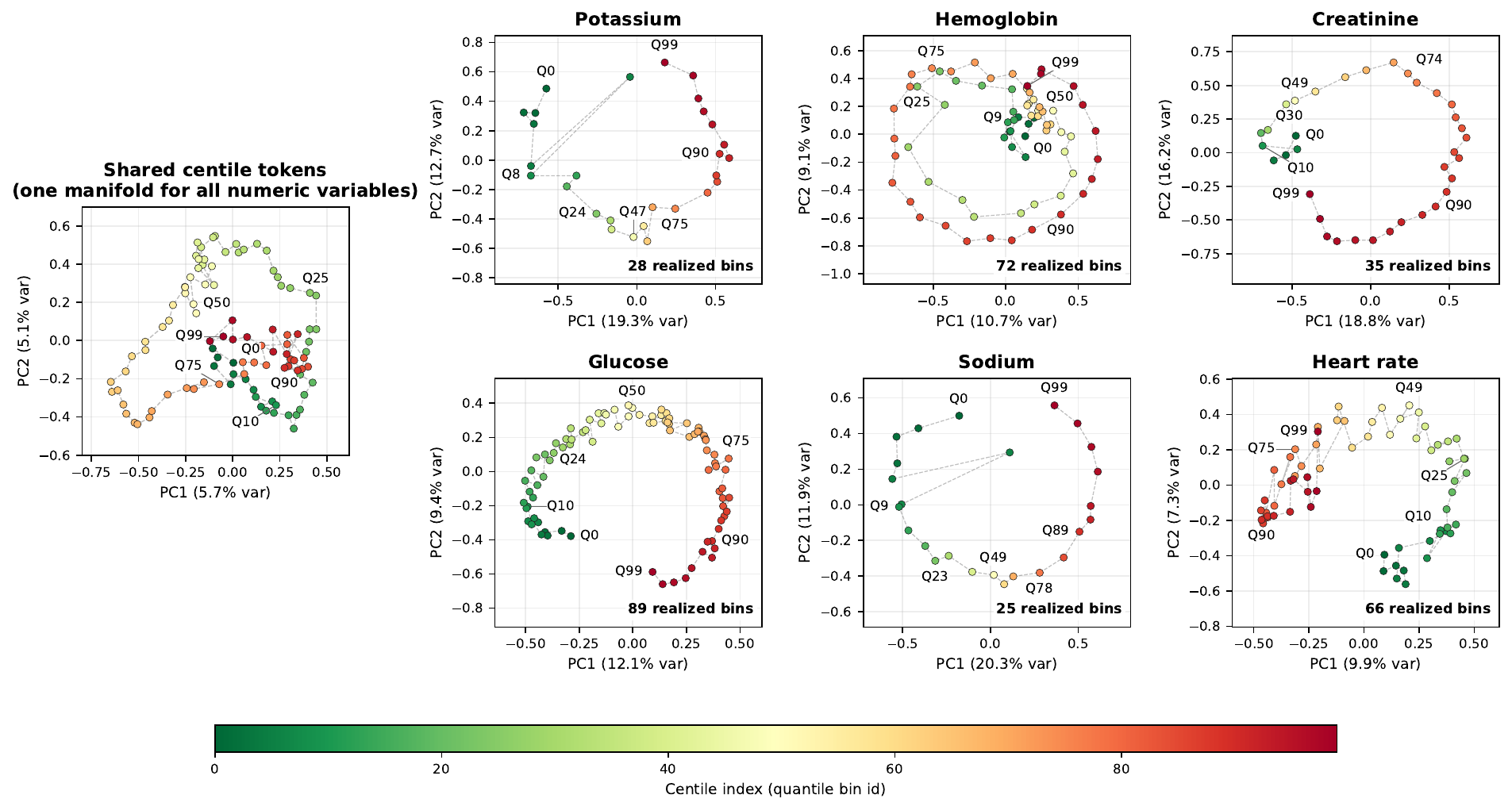}}
\end{figure*}

Cross-entropy training treats quantile bins as unrelated vocabulary entries with no direct supervision on numeric distance, yet the principal-component projections in \figureref{fig:pca_centile_geometry} recover clear one-dimensional ordering for both layouts. When all numeric measurements share 100 unfused centile tokens, their learned embeddings trace a single curve from \texttt{Q0} to \texttt{Q99}. Under fusion, where each measurement receives separate centile tokens across 25--89 realized bins, all six plotted variables still arrange into smooth ordered arcs along distinct code-specific trajectories.

\subsection{Linear Decodability of Clinical Boundaries}

\begin{figure*}[htbp]
\floatconts
  {fig:clinical_boundary_probe}
  {\caption{Clinical-boundary probe on unfused bin embeddings from the Llama-3.2-style decoder. Each population-quantile bin is labeled within or outside the MIMIC-IV reference interval from its midpoint; bars show leave-one-out logistic-regression accuracy, hollow diamonds mark the majority-class baseline (most frequent label per measurement and granularity), and BG denotes blood-gas measurements.}}
  {\includegraphics[width=0.72\linewidth]{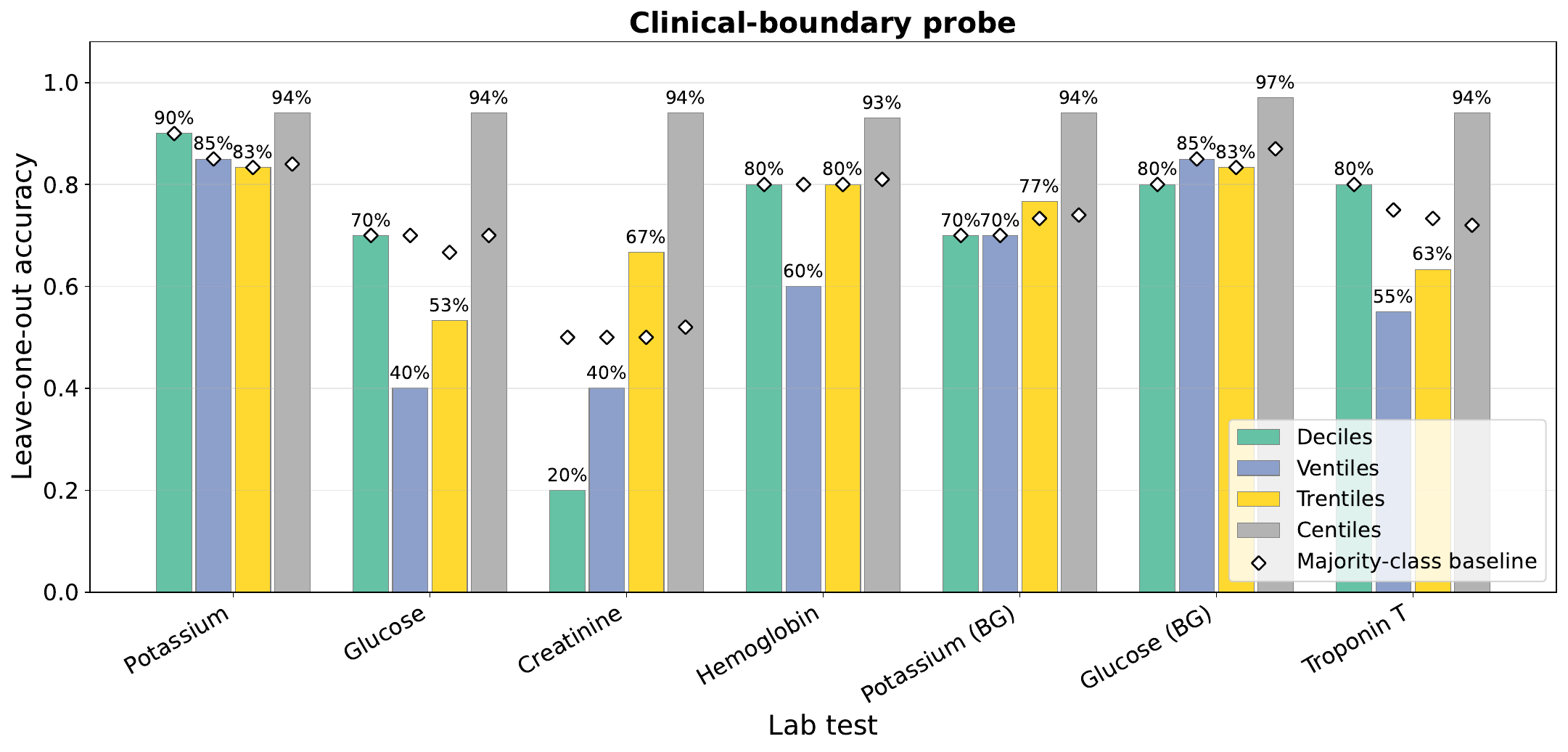}}
\end{figure*}

Population-quantile tokenizers set their bin cutoffs from training percentiles without access to clinical reference intervals. To check whether next-token training still separates normal from abnormal ranges linearly in embedding space, we labeled each population-quantile bin across seven laboratory tests and four granularities by comparing its midpoint against the MIMIC-IV reference interval and fit leave-one-out logistic regression to the unfused bin embeddings.

At deciles and ventiles, no probe outperformed its majority-class baseline, and at trentiles only creatinine and blood-gas potassium did (\figureref{fig:clinical_boundary_probe}). At centiles, however, all seven laboratory probes exceeded the majority-class baseline by 0.10--0.42, reaching 93--97\% leave-one-out accuracy, even though those finer bins did not improve downstream 24-hour probes in Experiment~1.
\FloatBarrier

\section{Outcome Definitions and Labeled Counts}
\label{app:appendix_results_tables}

\begingroup
\scriptsize
\setlength{\tabcolsep}{3pt}
\setlength{\LTleft}{0pt}
\setlength{\LTright}{0pt}
\begin{longtable}{@{}>{\raggedright\arraybackslash}p{0.28\textwidth}>{\raggedright\arraybackslash}p{\dimexpr\textwidth-0.28\textwidth-2\tabcolsep\relax}@{}}
  \caption{Definitions and MIMIC-IV source fields for $(24,48]$ follow-up outcomes, measured in hours from \texttt{hosp/\allowbreak admissions.\allowbreak admittime}. Hospital-status and ICU-stay outcomes use \texttt{deathtime}, \texttt{dod}, \texttt{dischtime}, \texttt{intime}, and \texttt{outtime}, whereas organ-support, laboratory, and vital-sign outcomes use \texttt{storetime} timestamps on \texttt{hosp/\allowbreak labevents}, \texttt{icu/\allowbreak chartevents}, \texttt{icu/\allowbreak inputevents}, and \texttt{icu/\allowbreak procedureevents}.
    Experiments~1--2 evaluate all 17 binary and 20 regression outcomes, while Experiment~3 evaluates 16 binary and 20 regression outcomes on the ICU cohort (omitting first ICU admission).}
  \label{tab:outcomes_sources}\\
  \toprule
  Outcome & Definition and source field(s) \\
  \midrule
  \endfirsthead
  \multicolumn{2}{@{}l}{\textbf{Table~\thetable\ continued from previous page}} \\
  \toprule
  Outcome & Definition and source field(s) \\
  \midrule
  \endhead
  \midrule
  \multicolumn{2}{r@{}}{\textit{Continued on next page}} \\
  \endfoot
  \bottomrule
  \endlastfoot
  \multicolumn{2}{@{}l}{\textbf{\textit{Binary outcomes (17)}}} \\*[2pt]
  Hospital mortality in hours 24--48
    & In-hospital death timestamp (\path{hosp/admissions.deathtime}, or \path{hosp/patients.dod} if null) within $[\texttt{admittime},\texttt{dischtime}]$ occurs in $(24,48]$, or \path{hosp/admissions.dischtime} occurs in $(24,48]$ with \path{hosp/admissions.discharge_location} equal to \texttt{DIED}. \\
  Still hospitalized at hour 48
    & Hospital discharge timestamp (\path{hosp/admissions.dischtime}) occurs after hour~48 relative to \path{hosp/admissions.admittime}. \\
  First ICU admission in hours 24--48
    & First ICU admission (\path{icu/icustays.intime}) occurs in $(24,48]$ after \path{hosp/admissions.admittime} (not in Experiment~3). \\
  Still in ICU at hour 48
    & An ICU stay interval from \path{icu/icustays.intime} to \path{icu/icustays.outtime} (\path{hosp/admissions.dischtime} if \texttt{outtime} is null) starts at ${\leq}48$~hours and ends after hour~48 (or has null \texttt{outtime}). \\
  Invasive mechanical ventilation
    & First invasive ventilation procedure (\path{icu/procedureevents.storetime}, \texttt{itemid} 224385 [intubation] or 225792 [invasive ventilation]) occurs in $(24,48]$. \\
  Vasopressor initiation
    & First vasoactive or inotropic infusion start (\path{icu/inputevents.storetime}, \texttt{itemid} 221906 [norepinephrine], 221289 or 229617 [epinephrine], 222315 [vasopressin], 221749 [phenylephrine], 221662 [dopamine], or 221653 [dobutamine]) occurs in $(24,48]$. \\
  Continuous renal replacement therapy initiation
    & First continuous renal replacement therapy event (\path{icu/chartevents.storetime} \texttt{itemid} 227290 or \path{icu/procedureevents.storetime} \texttt{itemid} 225802) occurs in $(24,48]$. \\
  Hemodialysis initiation
    & First hemodialysis event (\path{icu/chartevents.storetime} \texttt{itemid} 226499 or \path{icu/procedureevents.storetime} \texttt{itemid} 225441) occurs in $(24,48]$. \\
  Hyperkalemia
    & Maximum potassium in $(24,48]$ is ${\geq}6.5$~mEq/L (\path{hosp/labevents.valuenum}, \texttt{itemid} 50971 [chemistry] or 50822 [blood gas]). \\
  Severe hypokalemia
    & Minimum potassium in $(24,48]$ is ${<}2.5$~mEq/L (\path{hosp/labevents.valuenum}, \texttt{itemid} 50971 or 50822). \\
  Profound hyponatremia
    & Minimum sodium in $(24,48]$ is ${<}120$~mEq/L (\path{hosp/labevents.valuenum}, \texttt{itemid} 50983 [chemistry], 50824 [blood gas], or 52623 [chemistry, 0 events in MIMIC-IV]). \\
  Severe hypernatremia
    & Maximum sodium in $(24,48]$ is ${\geq}160$~mEq/L (\path{hosp/labevents.valuenum}, \texttt{itemid} 50983, 50824, or 52623). \\
  Hypoglycemia
    & Minimum glucose in $(24,48]$ is ${<}54$~mg/dL (\path{hosp/labevents.valuenum}, \texttt{itemid} 50931 [chemistry] or 50809 [blood gas]). \\
  Severe anemia
    & Minimum hemoglobin in $(24,48]$ is ${<}7.0$~g/dL (\path{hosp/labevents.valuenum}, \texttt{itemid} 51222 [hematology] or 50811 [blood gas]). \\
  Tachycardia
    & Maximum heart rate in $(24,48]$ is ${\geq}130$~bpm (\path{icu/chartevents.valuenum}, \texttt{itemid} 220045, values in $[20,300]$~bpm). \\
  Severe hypertension
    & Maximum systolic blood pressure is ${\geq}180$~mmHg (\texttt{itemid} 220050, 220179, or 225309, values in $[20,300]$~mmHg) or maximum diastolic blood pressure is ${\geq}120$~mmHg (\texttt{itemid} 220180, 220051, 225310, 224643, or 227242, values in $[20,300]$~mmHg) in $(24,48]$ from \path{icu/chartevents.valuenum}. \\
  Hypotension
    & Minimum mean arterial pressure is ${<}65$~mmHg (\texttt{itemid} 220052 or 220181, values in $[10,250]$~mmHg) or minimum systolic blood pressure is ${<}90$~mmHg (\texttt{itemid} 220050, 220179, or 225309, values in $[20,300]$~mmHg) in $(24,48]$ from \path{icu/chartevents.valuenum}. \\
  \midrule
  \multicolumn{2}{@{}l}{\textbf{\textit{Regression outcomes (20)}}} \\*[2pt]
  Min/max creatinine
    & Minimum and maximum creatinine (mg/dL) in $(24,48]$ from \path{hosp/labevents.valuenum} (\texttt{itemid} 50912). \\
  Min/max troponin
    & Minimum and maximum troponin (ng/mL) in $(24,48]$ from \path{hosp/labevents.valuenum} (\texttt{itemid} 51003 [Troponin~T] or 51002 [Troponin~I, 0 events in MIMIC-IV]). \\
  Min/max hemoglobin
    & Minimum and maximum hemoglobin (g/dL) in $(24,48]$ from \path{hosp/labevents.valuenum} (\texttt{itemid} 51222 [hematology] or 50811 [blood gas]). \\
  Min/max potassium
    & Minimum and maximum potassium (mEq/L) in $(24,48]$ from \path{hosp/labevents.valuenum} (\texttt{itemid} 50971 [chemistry] or 50822 [blood gas]). \\
  Min/max glucose
    & Minimum and maximum glucose (mg/dL) in $(24,48]$ from \path{hosp/labevents.valuenum} (\texttt{itemid} 50931 [chemistry] or 50809 [blood gas]). \\
  Min/max sodium
    & Minimum and maximum sodium (mEq/L) in $(24,48]$ from \path{hosp/labevents.valuenum} (\texttt{itemid} 50983 [chemistry], 50824 [blood gas], or 52623 [chemistry, 0 events in MIMIC-IV]). \\
  Min/max NT-proBNP
    & Minimum and maximum N-terminal pro-B-type natriuretic peptide (pg/mL) in $(24,48]$ from \path{hosp/labevents.valuenum} (\texttt{itemid} 50963). \\
  Min/max heart rate
    & Minimum and maximum heart rate (bpm) in $(24,48]$ from \path{icu/chartevents.valuenum} (\texttt{itemid} 220045, values in $[20,300]$~bpm). \\
  Min/max systolic blood pressure
    & Minimum and maximum systolic blood pressure (mmHg) in $(24,48]$ from \path{icu/chartevents.valuenum} (\texttt{itemid} 220050, 220179, or 225309, values in $[20,300]$~mmHg). \\
  Min/max diastolic blood pressure
    & Minimum and maximum diastolic blood pressure (mmHg) in $(24,48]$ from \path{icu/chartevents.valuenum} (\texttt{itemid} 220180, 220051, 225310, 224643, or 227242, values in $[20,300]$~mmHg). \\
\end{longtable}
\endgroup

\paragraph{Eligibility and leakage control.}
First-ICU-admission, organ-support initiation, and physiologic-threshold tasks exclude all hospitalizations meeting the endpoint at or before hour~24 ($\texttt{time} \leq \texttt{admittime} + 24$~hours). First-ICU-admission and organ-support tasks are labeled negative when no qualifying event is recorded in $(24,48]$, whereas physiologic-threshold and regression tasks require a qualifying measurement in $(24,48]$. Hospital mortality, hospital occupancy at hour~48, and ICU occupancy at hour~48 use all cohort hospitalizations. \tableref{tab:appendix_binary_outcome_descriptives} gives the binary labeled counts.

\begin{table}[htbp]
  \centering
  \caption{Labeled test-set counts for 24--48h binary outcomes.
    Experiments~1--2 use the hospitalization cohort (85{,}057 test hospitalizations).
    Experiment~3 uses the ICU cohort (16{,}529 test hospitalizations) and omits first ICU admission because every hospitalization in that cohort has an ICU stay.
    Positives are events observed in hours 24--48 among hospitalizations eligible for that label.}
  \label{tab:appendix_binary_outcome_descriptives}
  \scriptsize
  \setlength{\tabcolsep}{3pt}
  \begin{tabular}{p{0.28\textwidth}rrrr}
    \toprule
    Outcome & Exp.~1--2 labeled $N$ & Exp.~1--2 pos / neg & Exp.~3 labeled $N$ & Exp.~3 pos / neg \\
    \midrule
    Hospital mortality in hours 24--48 & 85,057 & 238 / 84,819 & 16,529 & 209 / 16,320 \\
    Still hospitalized at hour 48 & 85,057 & 66,238 / 18,819 & 16,529 & 15,438 / 1,091 \\
    First ICU admission in hours 24--48 & 71,949 & 1,004 / 70,945 & N/A & N/A \\
    Still in ICU at hour 48 & 85,057 & 8,018 / 77,039 & 16,529 & 8,041 / 8,488 \\
    Invasive mechanical ventilation & 83,278 & 1,153 / 82,125 & 14,766 & 1,224 / 13,542 \\
    Vasopressor initiation & 82,207 & 592 / 81,615 & 13,654 & 642 / 13,012 \\
    Continuous renal replacement therapy initiation & 84,947 & 56 / 84,891 & 16,429 & 71 / 16,358 \\
    Hemodialysis initiation & 84,866 & 118 / 84,748 & 16,330 & 135 / 16,195 \\
    Hyperkalemia & 58,971 & 158 / 58,813 & 14,635 & 99 / 14,536 \\
    Severe hypokalemia & 59,656 & 36 / 59,620 & 14,984 & 26 / 14,958 \\
    Profound hyponatremia & 59,350 & 29 / 59,321 & 14,858 & 17 / 14,841 \\
    Severe hypernatremia & 59,502 & 16 / 59,486 & 14,976 & 16 / 14,960 \\
    Hypoglycemia & 58,499 & 325 / 58,174 & 14,803 & 85 / 14,718 \\
    Severe anemia & 57,619 & 888 / 56,731 & 13,991 & 341 / 13,650 \\
    Tachycardia & 10,955 & 525 / 10,430 & 11,148 & 526 / 10,622 \\
    Severe hypertension & 10,580 & 783 / 9,797 & 10,740 & 784 / 9,956 \\
    Hypotension & 5,003 & 1,863 / 3,140 & 5,176 & 1,997 / 3,179 \\
    \bottomrule
  \end{tabular}
\end{table}

\FloatBarrier

\section{Cohort and Data Extraction}
\label{app:cohort}

We use MIMIC-IV v3.1 \citep{Johnson23MIMICIV,Johnson24MIMICIV31} (\tableref{tab:cohort_summary}) with patient-level train, validation, and test proportions of 70\%, 10\%, and 20\%. All hospitalizations from one patient stay in one partition. Experiments~1--2 include hospitalizations of at least 24 hours, which gives 85{,}057 test hospitalizations across 37{,}234 patients. In that test split, 63{,}897 hospitalizations (75.1\%) come from patients with two or more hospitalizations in that split. Experiment~3 restricts to hospitalizations of at least 24 hours with a linked \texttt{icu/icustays} record under a separate patient-level split (16{,}529 test hospitalizations).

Experiment~3 has more positives than Experiments~1--2 on some ICU endpoints because the two use separate patient partitions (\tableref{tab:appendix_binary_outcome_descriptives}). Invasive mechanical ventilation, for example, has 1{,}224 positives in Experiment~3 and 1{,}153 in Experiments~1--2. At hour 48, 8{,}041 hospitalizations remain in the ICU in Experiment~3 and 8{,}018 in Experiments~1--2.

\paragraph{Additional evaluation metrics.} For binary outcomes, we also compute Brier score and expected calibration error with 15 bins in addition to AUROC. For regression outcomes, we compute the coefficient of determination ($R^2$), mean absolute error, and root mean squared error next to Spearman correlation. Main-text comparisons use family-mean AUROC and Spearman correlation (\appendixref{app:result_tables}).

\begin{table}[htbp]
\floatconts
  {tab:cohort_summary}
  {\caption{Patients, hospital admissions, and ICU stays are MIMIC-IV v3.1 release totals \citep{Johnson24MIMICIV31}. Laboratory-event rows come from the MEDS extraction. The experimental cohorts are subsets of these totals.}}
  {\begin{tabular}{@{}lr@{}}
    \toprule
    Statistic & Count \\
    \midrule
    Patients & 364,627 \\
    Hospital admissions & 546,028 \\
    ICU stays & 94,458 \\
    Laboratory events & 84,133,368 \\
    Unique laboratory codes & 895 \\
    Codes with reference ranges & 334 \\
    \bottomrule
  \end{tabular}}
\end{table}

\paragraph{Data extraction.}
Experiments~1--2 use the MEDS extraction pipeline adopted from ETHOS-ARES \citep{Renc25ETHOSARES}, augmented with laboratory reference-range fields. We exclude administrative and billing code tables that can change after discharge to decrease temporal leakage \citep{Ramadan25Leakage}. For tables containing \texttt{storetime}, that field determines when information enters the event sequence. Experiment~3 uses the CLIF~2.1.0 conversion of MIMIC-IV \citep{Liao26MIMICIVExtCLIF}.
\FloatBarrier

\section{Training Details and Compute Environment}
\label{app:compute_by_stage}

\paragraph{Additional training details.} The learning rate warms up linearly over the first 1\% of optimizer updates. We clip gradient norms at 1.0 and use no gradient accumulation. Packed Experiment~1 runs have a $372{,}271$-step budget ($297{,}817$ training hospitalizations) and evaluate every $37{,}227$ updates. With unfused population deciles, the Llama-3.2-style decoder holds 123.6 million parameters (123{,}560{,}960) and the Qwen3-style decoder holds 103.7 million (103{,}734{,}272). Each architecture has 83.9 million non-embedding parameters. Parameter counts vary with vocabulary size because Llama-3.2-style decoders keep separate input and output embeddings, whereas Qwen3-style decoders tie them (\tableref{tab:exp1_vocab_params}).

Each model trained on a single NVIDIA RTX PRO 6000 or NVIDIA H200 GPU, and tokenization and linear probes ran on CPUs. Interrupted cluster runs resumed from saved checkpoints. Attention layers used FlashAttention-2 \citep{Dao24FlashAttention2} or PyTorch scaled dot-product attention.

\begin{table}[htbp]
\floatconts
  {tab:exp1_vocab_params}
  {\caption{Vocabulary sizes and total parameter counts (in millions) across the 12 Experiment~1 configurations. Non-embedding transformer layers have 83.9M parameters in each decoder. Llama-3.2-style decoders use untied input and output embeddings, whereas Qwen3-style decoders tie them.}}
  {\small
  \begin{tabular}{@{}lrrrrrr@{}}
    \toprule
    & \multicolumn{2}{c}{Vocabulary size} & \multicolumn{2}{c}{Llama-3.2-style (M)} & \multicolumn{2}{c@{}}{Qwen3-style (M)} \\
    \cmidrule(lr){2-3}\cmidrule(lr){4-5}\cmidrule(l){6-7}
    Quantization & Unfused & Fused & Unfused & Fused & Unfused & Fused \\
    \midrule
    Deciles (population) & 19,364 & 30,524 & 123.6 & 146.4 & 103.7 & 115.2 \\
    Ventiles (population) & 19,374 & 39,807 & 123.6 & 165.4 & 103.7 & 124.7 \\
    Ventiles (5--10--5 reference) & 19,374 & 38,837 & 123.6 & 163.4 & 103.7 & 123.7 \\
    Trentiles (population) & 19,384 & 47,519 & 123.6 & 181.2 & 103.8 & 132.6 \\
    Trentiles (10--10--10 reference) & 19,384 & 46,209 & 123.6 & 178.5 & 103.8 & 131.2 \\
    Centiles (population) & 19,454 & 83,937 & 123.7 & 255.8 & 103.8 & 169.9 \\
    \bottomrule
  \end{tabular}}
\end{table}

\paragraph{Experiment~3 training token exposure for native MIMIC versus CLIF.}
The full-hospitalization training split contained 684{,}688{,}603 native MIMIC tokens and 195{,}869{,}100 CLIF tokens before windowing (28.6\%). We partitioned those sequences into contiguous 4096-token windows (capped at 128 windows per hospitalization), which gave 195{,}380 native MIMIC and 82{,}372 CLIF training windows in each epoch. With one continuation token on each follow-on window and padding excluded, each epoch had 684{,}298{,}038 native MIMIC and 195{,}893{,}782 CLIF tokens. Each of the 12 runs completed five epochs, for 4{,}001{,}382{,}400 native or 1{,}686{,}978{,}560 CLIF padded positions. Tokenizer vocabulary sizes were 17{,}752 for native MIMIC and 397 for CLIF. Corresponding parameter counts were 120.3M and 84.7M on Llama-3.2-style decoders, and 102.1M and 84.3M on Qwen3-style decoders.
\FloatBarrier

\section{Tokenized Sequence Length Distributions}
\label{app:length_distributions}

In Figures~\ref{fig:seq_len_distribution} and~\ref{fig:temporal_len_distribution}, we plot training sequence lengths across fusion and temporal-encoding choices for Experiments~1--2. Experiment~3 fits separate tokenizers on the ICU cohort (\appendixref{app:compute_by_stage}, \appendixref{app:numeric_streams}).

\begin{figure}[htbp]
\floatconts
  {fig:seq_len_distribution}
  {\caption{Token lengths for the fused and unfused tokenizers in Experiments~1--2. We compute histograms from token sequences of training hospitalizations before context-window truncation, for full and first-24h event streams. Dashed lines mark 1024, 2048, and 4096 tokens. More than 99.95\% of first-24h training hospitalizations have length at most 4096 tokens.}}
  {\includegraphics[width=\linewidth]{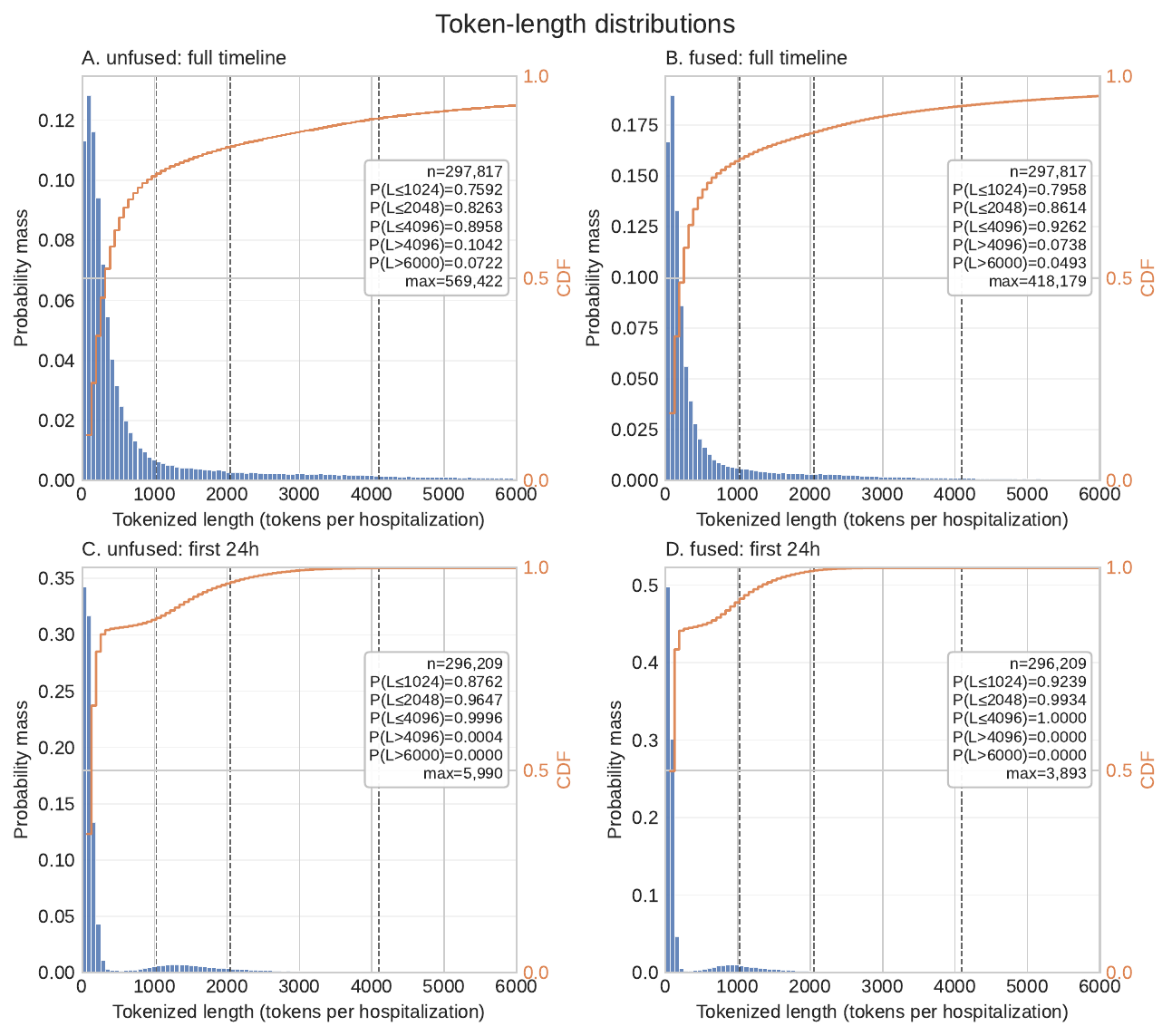}}
\end{figure}

\begin{figure}[htbp]
\floatconts
  {fig:temporal_len_distribution}
  {\caption{Token lengths with and without explicit time tokens for the tokenizers in Experiments~1--2. Omitting time tokens decreases median first-24h length from 93 to 83 tokens, approximately 11\%. Event order and admission-relative RoPE use the same token sequence and differ only in their rotary position indices (sequential token index versus 30-second steps since admission).}}
  {\includegraphics[width=\linewidth]{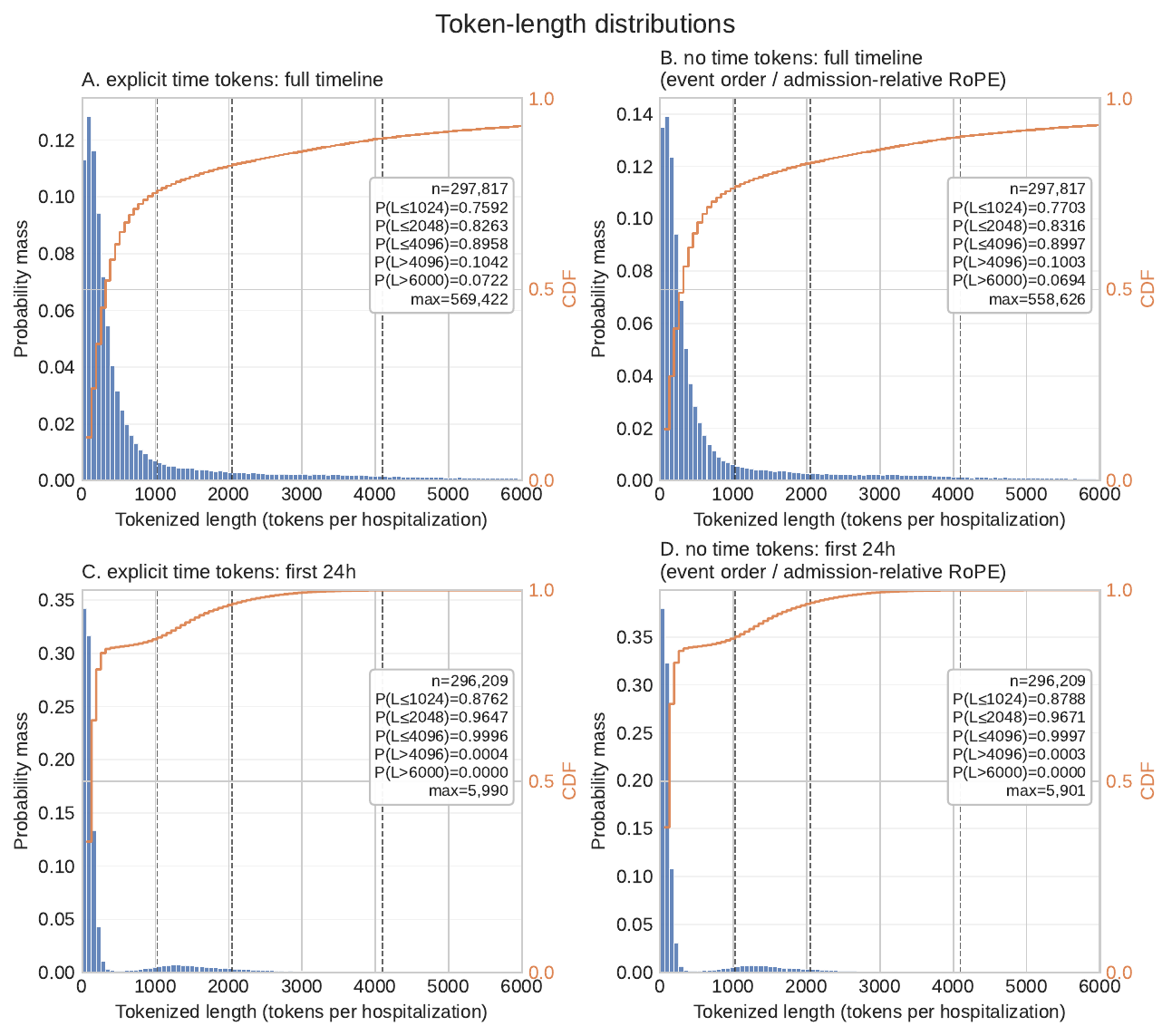}}
\end{figure}

\paragraph{Packed-sequence boundaries.}
Experiment~1 concatenates hospitalizations into fixed-length training blocks. A short, Poisson-distributed run of padding tokens (mean 7) separates adjacent hospitalizations. This padding limits direct continuity across boundaries and exposes the model to padding tokens during training \citep{Burkhart25FoundationModels,Burkhart26QuantifyingSurprise}.
\FloatBarrier

\section{Formal Definitions of Value Encoders}
\label{app:value_encoders_formal}

\paragraph{Soft discretization.}
For $K$ quantile bins with interior cutoffs $b_0\leq\dots\leq b_{K-2}$, let $b_{k-1}<v\leq b_k$ ($1\leq k\leq K-2$) be the adjacent cutoffs surrounding $v$, and define
\[
\alpha=\frac{v-b_{k-1}}{b_k-b_{k-1}},
\]
with $\alpha=0$ when the boundaries are equal. A value equal to boundary $b_k$ stays in bin $k$, consistent with the half-open intervals $(b_{k-1}, b_k]$ of the discrete tokenizer. At a quantile-token target, the same weight defines
\[
\mathcal{L}_{\mathrm{soft}}(v)
=-(1-\alpha)\log p(Q_{k-1}\mid\cdot)-\alpha\log p(Q_k\mid\cdot).
\]
All other targets keep standard causal cross-entropy. Values outside the interior cutoff range $(b_0, b_{K-2}]$ use the first or last bin embedding.

\paragraph{Code-normalized xVal.}
For numeric value $v$ from code $c$, let $m_c$ be the code-specific training median. The robust scale is
\[
s_c=\max(\operatorname{IQR}_c/1.35,10^{-8}).
\]
The normalized value is
\[
z=\operatorname{clip}\left(\frac{v-m_c}{s_c}, -5, 5\right).
\]
The auxiliary head predicts $z$ at numeric positions. If a code has no numeric values in the training split, numeric scaling and the auxiliary loss are omitted for that event.

\paragraph{Affine xVal.}
Affine xVal uses the same $z$ and auxiliary head. Its embedding is
\[
\mathbf{e}(v)=z\mathbf{e}_{\texttt{[NUM]}}+\mathbf{b}.
\]
Here, $\mathbf{e}_{\texttt{[NUM]}}$ is the shared numeric-placeholder embedding and $\mathbf{b}$ is a learned bias vector. The code-median value thus has embedding $\mathbf{b}$ rather than the zero vector.

\section{Numeric Event Families}
\label{app:numeric_streams}

In Experiments~1--2, the tokenizer quantizes laboratory results, vital signs,
infusion start rates, patient weight at infusion, fluid outputs, and admission
age. Their event sources are \path{hosp/labevents.valuenum},
\path{icu/chartevents.valuenum}, \path{icu/inputevents.rate},
\path{icu/inputevents.patientweight}, and \path{icu/outputevents.value}.
Medication, infusion-end, transfer, ICU admission/discharge, and procedure
events stay categorical in Experiments~1--2, even when MEDS records an infusion
amount or procedure duration. Reference-range anchoring applies only to
laboratory events with valid lower and upper reference fields. Other numeric
families use population quantiles.

In Experiment~3, we first convert CLIF~2.1.0 tables into MEDS format. We then
fit separate unfused-decile tokenizers on the native MIMIC and CLIF ICU
training splits under one MEDS configuration. Here, we quantize all
non-admission and non-discharge events with a non-null \texttt{numeric\_value}.
Thus, native MIMIC additionally quantizes infusion-end and procedure values, and CLIF
quantizes numeric laboratory, vital-sign, assessment, respiratory, medication,
and CRRT events.

\section{Statistical Comparisons}
\label{app:statistical-comparisons}
Our main tables report scores for eight outcome families across two architectures and three initialization seeds. Within each model, outcomes in a family are equally weighted, and the six model scores are averaged to obtain the family mean. For each comparison, we pair the alternative model with its baseline by architecture and seed ($n=6$ pairs), defining $d_m$ as the unrounded alternative-minus-baseline score for pair $m$ with mean $\Delta$ and standard error $\mathrm{SE}$. Because family and outcome $\Delta$ values are computed before three-decimal rounding, a reported $\Delta$ can differ by $0.001$ from the difference between rounded six-model means. We use a two-sided one-sample $t$-test ($t = \Delta / \mathrm{SE}$, $\mathrm{df}=5$) and report 95\% $t$-intervals $\Delta \pm 2.571\,\mathrm{SE}$ (bounds shown as $0.000$ also result from three-decimal rounding).

The baselines are unfused deciles for quantization, matched unfused bins for fusion, discrete deciles with time tokens for Experiment~2, and native MIMIC for CLIF. Each comparison group multiplies its $\Delta$ columns in Tables~\ref{tab:exp1_quantization}--\ref{tab:exp3_family} by the eight outcome families, yielding $17\times 8=136$ family comparisons overall. We apply Benjamini--Hochberg (BH) correction at $q<0.05$ within quantization ($5\times 8=40$ tests, \tableref{tab:exp1_quantization}), fusion ($6\times 8=48$, \tableref{tab:exp1_fusion}), numeric encoding ($3\times 8=24$, \tableref{tab:exp2_numeric}), temporal encoding ($2\times 8=16$, \tableref{tab:exp2_numeric}), and schema remapping ($1\times 8=8$, \tableref{tab:exp3_family}) \citep{Benjamini95FDR}. Across the 136 family comparisons, 74 are significant at unadjusted $p<0.05$ and 69 remain significant after BH adjustment. Outcome-level comparisons use the same six-pair $t$-test with BH correction within each configuration comparison. To assess decoder sensitivity, we use a two-sided Welch $t$-test to compare the three Llama-3.2-style and three Qwen3-style differences under the same BH correction (0 of 136 significant, smallest BH $p=0.56$).

The six-pair $t$-test captures replication uncertainty across architectures and random seeds on the fixed test set. To estimate test-cohort sampling uncertainty under multi-stay correlation, we run a 1{,}000-draw patient-clustered bootstrap (seed 42). In each draw, we resample test patients with replacement, retain all eligible hospitalizations for each sampled patient, and recompute the six-model $\Delta$. Across the 136 family comparisons, 90 patient-clustered 95\% percentile intervals do not include zero, including all 69 BH-significant comparisons (38 fusion, 16 numeric, 9 temporal, and 6 CLIF).

\section{Family-Mean and Outcome-Level Summaries}
\label{app:result_tables}

For each model, we average metrics across outcomes within the family and then average across the two architectures and three initialization seeds, reporting the six-model mean and min/max range in brackets (\tableref{tab:appendix_family_means_exp1}, \tableref{tab:appendix_family_means_exp2}, \tableref{tab:appendix_family_means_exp3}). Full outcome-level tables (\tableref{tab:appendix_outcome_means_exp1}, \tableref{tab:appendix_outcome_means_exp2}, \tableref{tab:appendix_outcome_means_exp3}) list the six-model mean, minimum, and maximum for every outcome to show variation across trained models.

In \texttt{input-representation-benchmark}, \path{outputs/runs/ml4h_2026/metrics/metrics_long.parquet} stores run-level AUROC, Brier score, 15-bin expected calibration error, Spearman $\rho$, $R^2$, mean absolute error, and root mean squared error (\appendixref{app:cohort}). Paired $t$-tests and Llama-versus-Qwen Welch $t$-tests (\appendixref{app:statistical-comparisons}) appear in \path{outputs/runs/ml4h_2026/model_level_tests/family_tests.csv} and \path{outputs/runs/ml4h_2026/model_level_tests/outcome_tests.csv}. Patient-clustered bootstrap intervals for the 136 family differences appear in \path{outputs/runs/ml4h_2026/patient_cluster_bootstrap/patient_cluster_intervals.parquet}.

\begin{table}[htbp]
\floatconts
  {tab:appendix_family_means_exp1}
  {\caption{Family-mean AUROC and Spearman~$\rho$ for every Experiment~1 configuration. Each cell reports the mean across six trained models (three seeds $\times$ two architectures), with min/max in brackets. Member outcomes receive equal weight within each model before the six-model summary.}}
  {\resizebox{\linewidth}{!}{%
  \scriptsize
  \setlength{\tabcolsep}{3pt}

\endgroup

\begin{table}[htbp]
\floatconts
  {tab:appendix_family_means_exp2}
  {\caption{Family-mean AUROC and Spearman~$\rho$ for every Experiment~2 configuration. Each cell reports the mean across six trained models (three seeds $\times$ two architectures), with min/max in brackets. Member outcomes receive equal weight within each model before the six-model summary.}}
  {\resizebox{\linewidth}{!}{%
  \scriptsize
  \setlength{\tabcolsep}{3pt}
  %
\endgroup

\begin{table}[htbp]
\floatconts
  {tab:appendix_family_means_exp3}
  {\caption{Family-mean AUROC and Spearman~$\rho$ for every Experiment~3 configuration. Each cell reports the mean across six trained models (three seeds $\times$ two architectures), with min/max in brackets. Member outcomes receive equal weight within each model before the six-model summary.}}
  {\resizebox{\linewidth}{!}{%
  \scriptsize
  \setlength{\tabcolsep}{3pt}
  %
\endgroup

\FloatBarrier

\end{document}